\documentclass[runningheads]{llncs}

\usepackage{eccv}

\usepackage{eccvabbrv}

\usepackage{graphicx}
\usepackage{booktabs}

\usepackage[accsupp]{axessibility}  

\usepackage{hyperref}

\usepackage{orcidlink}

\begin{document}

\title{ACDG-VTON: Accurate and Contained Diffusion Generation for Virtual Try-On}

\titlerunning{ACDG-VTON}

\author{Jeffrey Zhang\inst{1,2} \and
Kedan Li\inst{1,2} \and
Shao-Yu Chang\inst{2} \and
David Forsyth\inst{2}
}

\authorrunning{J.~Zhang et al.}

\institute{Revery AI Inc.\\
\email{\{jeff, kedan\}@revery.ai}\\
\and
University of Illinois, Urbana-Champaign\\
\email{\{shaoyu3, daf\}@illinois.edu}}

\maketitle

\begin{abstract}
Virtual Try-on (VTON) involves generating images of a person wearing selected garments. Diffusion-based methods, in particular, can create high-quality images, but they struggle to maintain the identities of the input garments. We identified this problem stems from the specifics in the training formulation for diffusion. To address this, we propose a unique training scheme that limits the scope in which diffusion is trained. We use a control image that perfectly aligns with the target image during training. In turn, this accurately preserves garment details during inference. We demonstrate our method not only effectively conserves garment details but also allows for layering, styling, and shoe try-on. Our method runs multi-garment try-on in a single inference cycle and can support high-quality zoomed-in generations without training in higher resolutions. Finally, we show our method surpasses prior methods in accuracy and quality.

\end{abstract}

\section{Introduction}
\label{sec:intro}

\begin{figure}[tb]
  \centering
    \includegraphics[width=1.0\textwidth]{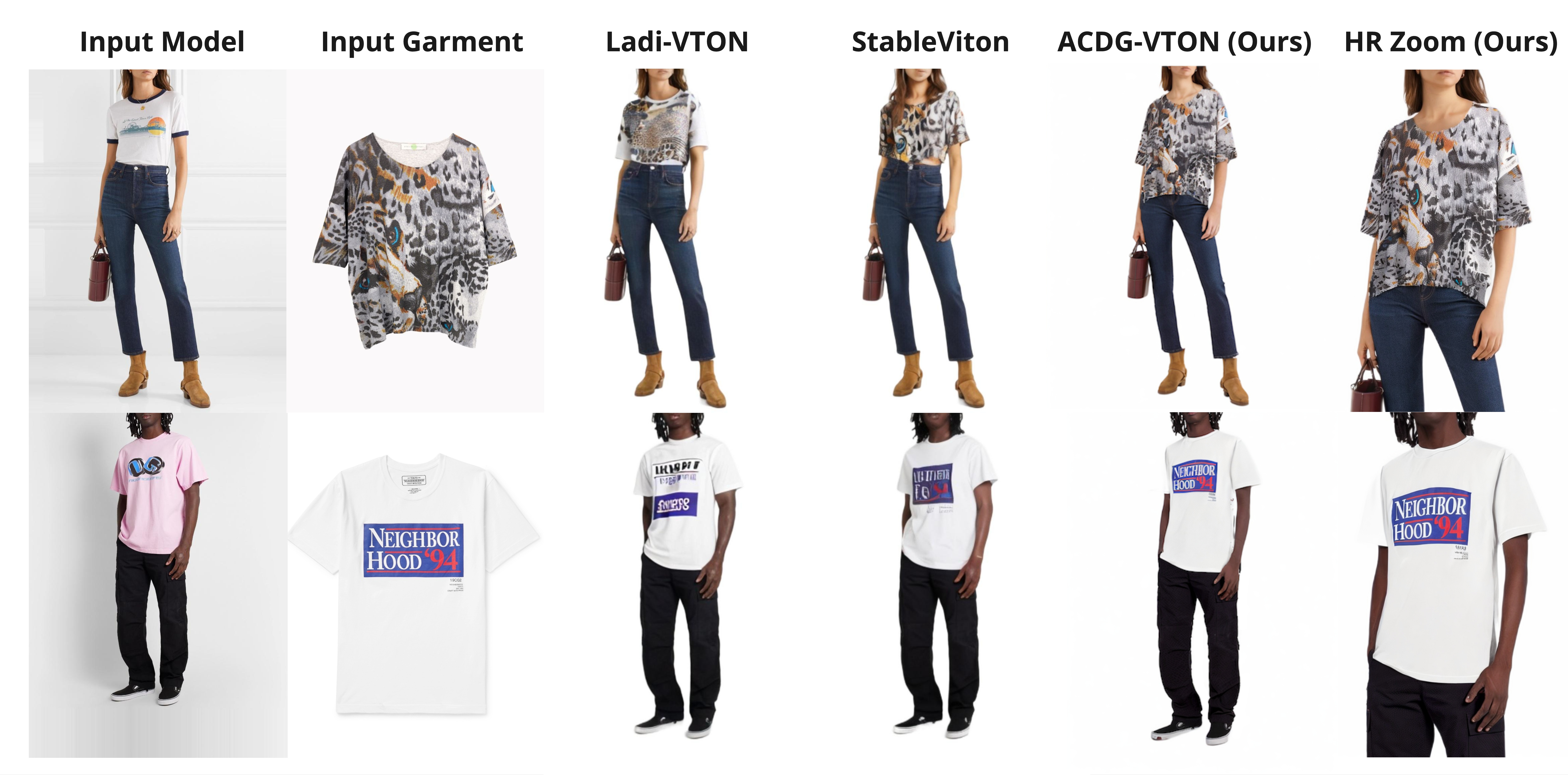}
  \caption{Accuracy refers to how well the generated items generate the details of the actual garments. Our method preserves garment details such as graphics, text, and patterns better than diffusion-based try-on methods. LaDI-VTON~\cite{Morelli2023LadiVTON} and StableVITON~\cite{Kim2023StableVITON} are not accurate because both methods alter and hallucinate details on the garment. Furthermore, our proposed HR Zoom method can make additional improvements to preserve the details that VAEs may distort (e.g. text and small patterns).}
  \label{fig:teaser}
\end{figure}

Virtual Try-on (VTON) is hard for modern generative models because VTON methods, to be commercially
successful, must actually show a {\em specific} garment {\em accurately}. Achieving this criterion
is a demanding task (Fig.~\ref{fig:teaser} and~\ref{fig:sota_comparison}). Together with accuracy,
VTON pipelines must also offer quality and controllability to be commercially appealing. At present, warp-then-GAN
pipelines yield the most accurate and controllable methods, but often produce images
with quality issues. Conversely, diffusion methods produce high quality images, but often
hallucinate non-existent garment features.

We describe a novel warp-then-diffuse method, ACDG-VTON, that accurately reproduces garment details,
notably improves image quality, and allows for garment controllability. Qualitative results show ACDG-VTON preserves garment features (text, logos, textures, and patterns) better than other diffusion-based methods and outperforms them in standard
quantitative metrics. Similarly, user studies show a pronounced preference for our method in accuracy. Finally, user studies also show a significant gain in quality from
applying properly contained diffusion as a generator compared to GAN approaches.  

VTON systems are intended to sell garments, and the application requires
showing  {\em specific} garments {\bf accurately}, {\bf controllably}, and at high {\bf quality}.
\textbf{Accuracy} refers to how well the generated items mirror the
characteristics of the actual garments in (say) color, shape, texture, logos, proportions, and other
aspects.  For a commercial system to be successful, it must be accurate because a consumer who is
misled by a VTON system may return purchased garments. \textbf{Quality} refers to the realism and
visual appeal of generated images. Quality is important in commercial systems because consumers who
find images unattractive or "flat" might not purchase garments. \textbf{Controllability} allows,
for example, a user to layer garments differently, choose styling options, choose model poses, etc. A controllable system offers consumers an engaging experience and vendors the opportunity to style garments differently depending on what outfit a garment appears in.

\begin{figure}[tb]
  \centering
    \includegraphics[width=0.98\textwidth]{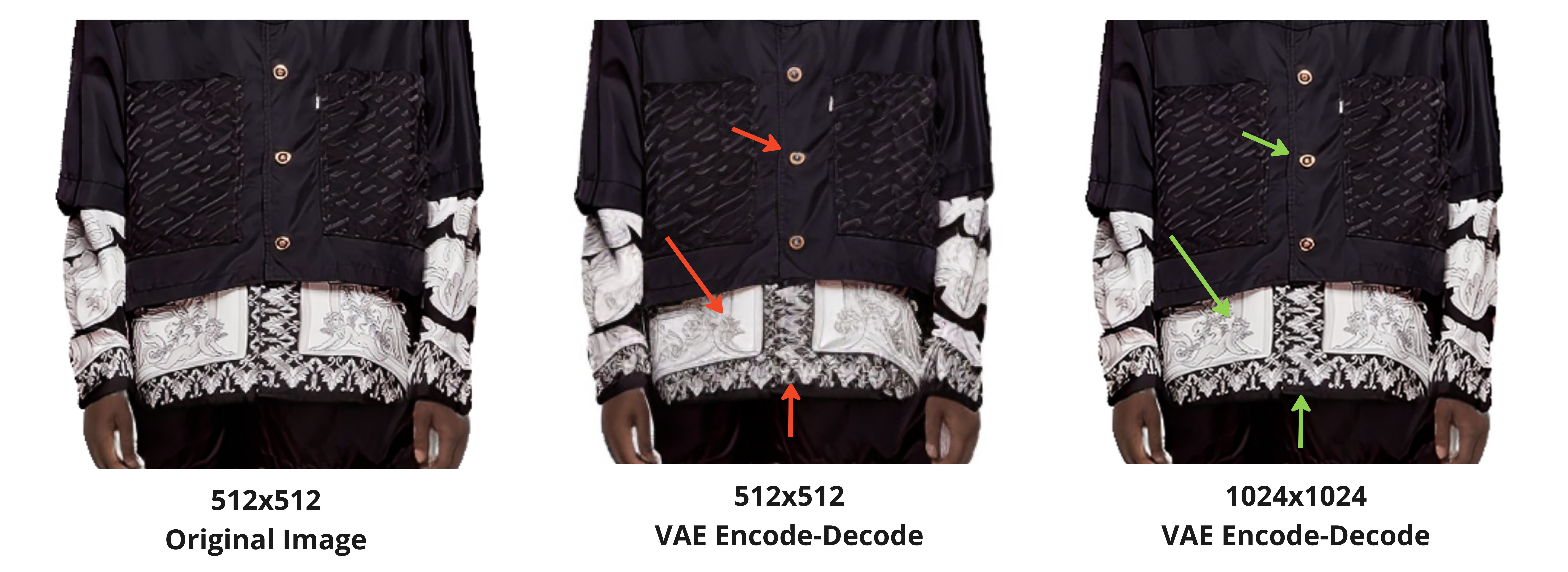}
  \caption{VAEs~\cite{Kingma2014} can cause loss of high-frequency details. If we take a 512x512 crop from a larger image and encode and decode the image through a VAE, certain details will be altered because the VAE dictionary cannot reconstruct high-frequency features. The button's colors are changed, the angel's face and body are altered, and the pattern near the hem differs (see red arrows). We can work around this issue by upsampling the 512x512 crop to 1024x1024. The details are better preserved if we encode and decode the 1024x1024 image through VAE (see green arrows). 
  }
  \label{fig:vae_encode_decode}
\end{figure}

Accuracy problems with diffusion-based methods~\cite{Morelli2023LadiVTON, Kim2023StableVITON, Gou2023DCIVTON,
Xie2023GPVTON} have two main sources.  First, diffusion methods tend to hallucinate details given an opportunity.  For example, in Fig.~\ref{fig:teaser}, diffusion
methods like LaDI-VTON~\cite{Morelli2023LadiVTON} and StableVITON~\cite{Kim2023StableVITON} create
designs that were not included in the input garment. Second, most diffusion methods use a Variational Autoencoder (VAE)~\cite{Kingma2014} for
efficiency. As Fig.~\ref{fig:vae_encode_decode} shows, the VAE encoding may distort image details because the size of 
the VAE dictionary is limited, and leading to problems with fine details. 

Our system is carefully engineered to maintain the accuracy and controllability of warp-then-GAN pipelines,
while using diffusion generation to improve the quality. At inference, our approach warps garments to create
a control image that is fed to a diffusion denoiser, which produces the final image.
The key innovation is preventing hallucinations by carefully choosing the control images used in training
the diffusion models.  Rather than using warped garment images, we craft a control image that is precisely aligned with
the target image to limit the scope offered to the diffusion model, leading to an improvement in accuracy. Additionally, our approach has added benefits of garment controllability and the ability to generate high-resolution zoomed (HR Zoom) close-up images of VTON.

\section{Related Works}
\label{sec:related}
\subsection{Image-based Virtual Try-on}
The main challenge of image-based VTON is producing high-quality images while faithfully preserving the garment's identity. Modern image generation networks trained with adversaries can produce VTON images but have difficulty preserving the exact geometrical patterns (such as logos, prints, etc..) on the garment. As a result, most state-of-the-art VTON pipelines learn a differentiable warper to align the garment onto the person, thus preserving the geometrical patterns and details~\cite{Dong2018SoftGatedWF, Lassner:GeneratingPeople:2017, hsiao2019fashionplus, zhu2017be,Han_2019_ICCV, wang2018toward, Rocco17, tprvton, Issenhuth2020DoNM, ge2021parser, cycle_consistency_tryon, choi2021vitonhd, Chopra_2021_ICCV, Style_Based_Global_Appearance, Fele_2022_WACV, morelli2022dresscode, lee2022hrviton, Bai2022SingleSV, Kedan_Li_2021_CVPR, Xie2023GPVTON, Li2023POVNet}. Image warping processes apply a spatial transformation to an image and thus can preserve the 2D patterns on a garment. The warped garment images are provided to an image generator network to synthesize an image of a person wearing the garment. Furthermore, \cite{Li2024Controlling, Chen2023Size} show that using explicit warps allows changing the size of the garments and adjusting how they drape on the model. 
 
Diffusion-based try-on methods~\cite{Zhu2023TryOnDiffusion, Chen2023AnyDoor, Gou2023DCIVTON, Morelli2023LadiVTON, Kim2023StableVITON} have shown the ability to generate high-quality images. However, most methods use a VAE to reduce computational complexity, which limits the accuracy of the try-on to what the VAE can reconstruct. This encoding and decoding process leads to fine details being altered. While~\cite{Zhu2023TryOnDiffusion} is trained in the image space, it still hallucinates features and can alter the garment identity. ~\cite{Morelli2023LadiVTON, Gou2023DCIVTON} attempt to use warps as a reference to mitigate changes in garment identity, whereas~\cite{Kim2023StableVITON} tries to learn implicit warping without needing an external warping network. These methods all still hallucinate and alter garment details. Finally, these methods often rely on the diffusion network to decide how a garment is styled and generated, giving users little control over how the garment is worn.

We develop a diffusion-based method that addresses the main drawback of diffusion-based try-on, the ability to preserve details. Our method faithfully copies details from an explicit garment warp and allows control over multi-garment layering, styling, and shoe try-on. 
Furthermore, our proposed method can overcome the alterations to garment details by aligning our control image with the target image during training, and it can overcome VAE limitations by zooming in to generate high-frequency regions.

\subsection{Multi-Garment Virtual Try-on}
Multi-garment virtual try-on is more challenging than the single garment version because the framework needs to manage the layering between multiple garments and the body appropriately. O-VITON~\cite{Neuberger_2020_CVPR} first synthesizes a semantic layout to outline the garment interactions and then broadcasts the feature encoding vectors based on the layout. This formulation can manage the interactions between garments well. However, feature encoding vectors cannot preserve structural patterns, resulting in loss of details during rendering. Meanwhile, OVNet~\cite{Kedan_Li_2021_CVPR} proposed an iterative method of constructing the outfit, swapping one garment at each generation step. This framework is able to preserve the attributes but does not have a solution to coordinate the garment interactions in an outfit. In contrast, \cite{Li2024Controlling} applies procedural edits to predicted control points to improve layered outfits' rendering significantly. On the other hand, diffusion methods~\cite{Zhu2023TryOnDiffusion, Chen2023AnyDoor, Gou2023DCIVTON, Morelli2023LadiVTON, Kim2023StableVITON} are designed only for one garment at a time (often only trained on tops), and they do not provide reliable ways to control the interaction between garments. 

In our work, we show a multi-garment VTON method using diffusion, adapting the accuracy and controllability of \cite{Li2024Controlling} while notably improving the quality.

\section{Background VTON Notation}
The formulation for virtual try-on takes a garment image $g$ and a person model image $m$ to generate a final image $m^g$ where the person image $m$ is wearing the garment $g$. In many VTON pipelines~\cite{tprvton,Issenhuth2020DoNM, Kedan_Li_2021_CVPR, Chopra_2021_ICCV, Li2024Controlling}, a semantic layout generator $H$ is used to predict the post-tryon segmentation map of the garment on the model, and a warper $W$ is used to warp the garment $g$ to the approximate location on the model. The warper $W$ takes a garment $g$ and other features of the model (e.g. pose, parsing, etc.) to create a warped garment $g^w$. The semantic layout generator $H$ takes in $g$, $g^w$, and other person features (e.g. pose, parsing, etc.) to output a semantic segmentation parsing $p$ of the garment on the model post-tryon. Finally, an image generator takes features such as the warped garment $g^w$, the segmentation map $p$, the model image $m$, and other features to produce the final output image $\hat{m^g}$.

\section{Method}
\label{sec:method}
Our method, ACDG-VTON, is a diffusion-based pipeline that faithfully preserves garment details. We replace the standard VTON image generator with a diffusion denoiser $F$. We develop a novel training scheme that properly limits the scope of the diffusion training to improve accuracy, quality, and garment controllability while allowing for zoom-in generations.

ACDG-VTON comprises of two stages: 1) composing an incomplete image as a control image and starting point for diffusion and 2) running diffusion on the incomplete image to generate the final try-on output. A necessary feature of the control image is that it is aligned with the ground truth image during training. Without this alignment, diffusion produces inaccurate results (Sec.~\ref{sec:ablations}).

\begin{figure}[tb]
  \centering
    \includegraphics[width=0.88\textwidth]{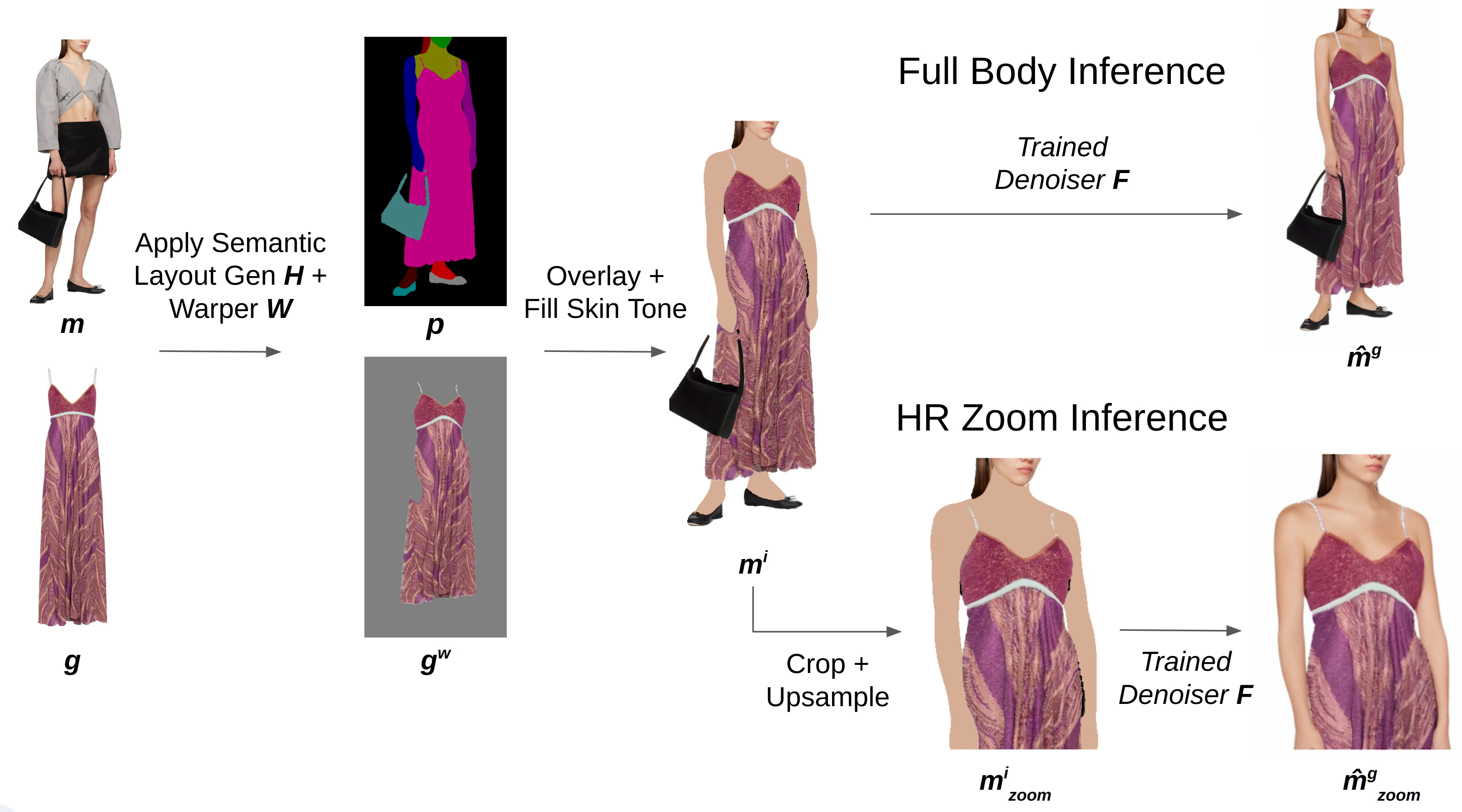}
  \caption{\textbf{Inference procedure}. We want model image $m$ to wear garment $g$. We apply the Semantic Layout Gen $H$ and Warper $W$ to get the semantic layout parsing $p$ and warped garment $g^w$ (we use $W$ and $G$ from ~\cite{Li2024Controlling}, but expect others (e.g.,~\cite{tprvton,Issenhuth2020DoNM, Kedan_Li_2021_CVPR, Chopra_2021_ICCV}) would work as well). We overlay warped garment $g^w$ on model image $m$, and we use $p$ to fill in the skin with the median pixel value from the face to get our incomplete image $m^i$ during inference (the control image is different for training; see Fig.~\ref{fig:warp_training}). For full body inference, we pass $m^i$ as the initialization, control, and image CLIP embedding to our trained diffusion model $F$ to get the try-on output $\hat{m}^g$ (architecture in Figure~\ref{fig:training_architecture}). 
  Notice our trained denoiser $F$ preserves garment details while fixing bad strap segmentations on the dress and the bag. For HR Zoom inference, we crop and upsample $m^i$ to create $m^i_{zoom}$. We run inference through the same trained denoiser $F$ to generate a close-up generation $\hat{m}^g_{zoom}$ that has the same resolution as $\hat{m}^g$.}
  \label{fig:diffusion_inference}
\end{figure}

\begin{figure}[tb]
  \centering
    \includegraphics[width=0.9\textwidth]{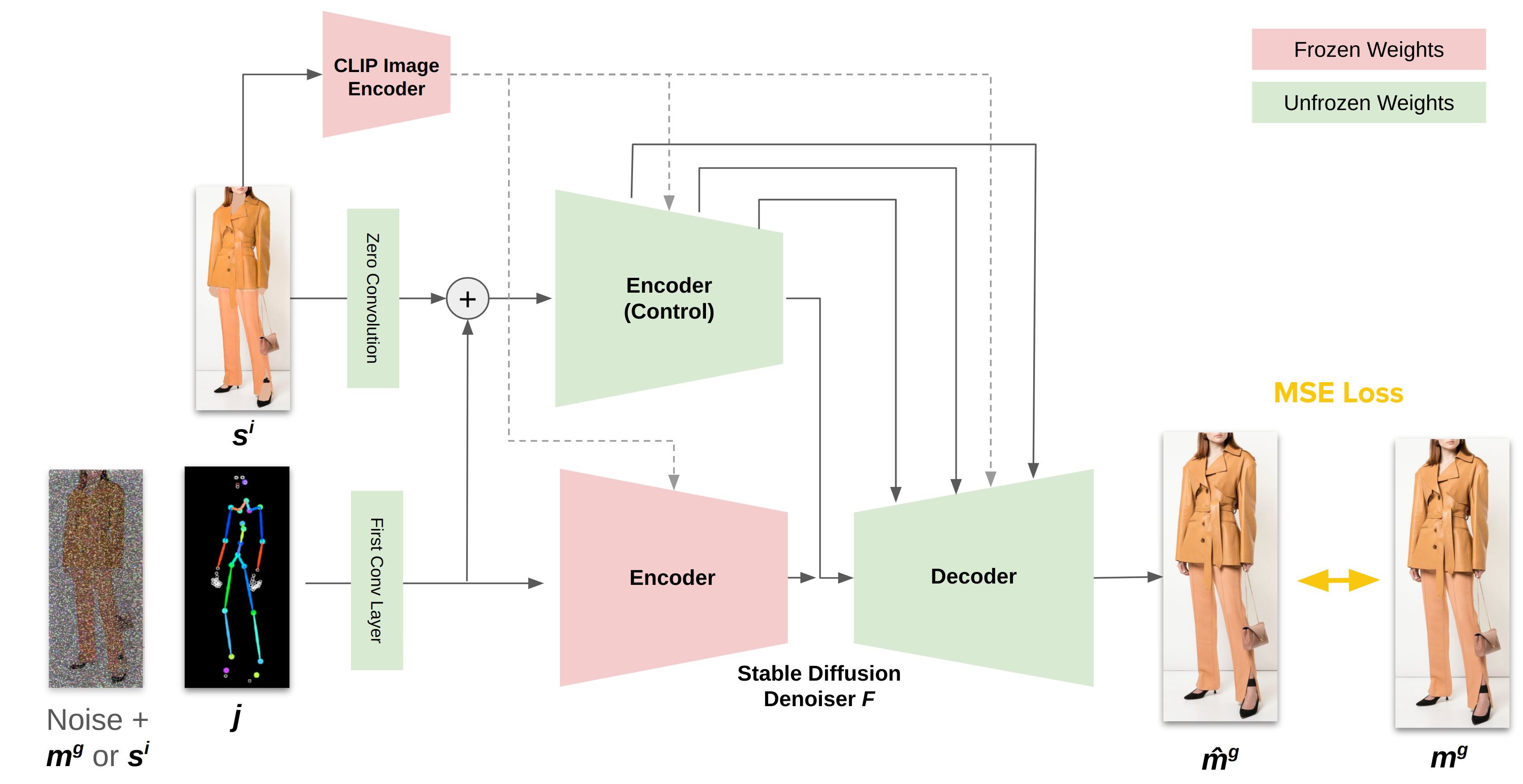}
  \caption{We adopt ControlNet~\cite{Zhang2023ControlNet} architecture to take in a simulated incomplete model image $s^i$ as a control to help preserve the identities of the garments. In addition, the control $s^i$ is used as the noisy image for the first few timesteps during training, and the ground truth model $m^g$ is used for the noisy image otherwise (Sec.~\ref{sec:training_method}, "Control Initialization"). An smplx joint image $j$ is concatenated to the noisy image as input to the diffusion network to better control the generation of arms and fingers. The control image $s^i$ is fed into a CLIP image encoder~\cite{ramesh2022} and used as the embedding condition for the diffusion network. Finally, an MSE loss is applied to the predicted noise compared to the added noise (we show images $\hat{m}^g$ and $m^g$ in the figure for simplicity).
  }
  \label{fig:training_architecture}
\end{figure}
\subsection{Inference Procedure}
During inference, we want to produce an image of the model depicted in image $m$ wearing some garment(s) $g$;  call this inference result $\hat{m}^g$. We obtain $\hat{m}^g$ by producing an incomplete model image $m^i$, then refining it with a diffusion-based generator $F$. The incomplete model image is created with warped garments overlaid on $m$ with skin filled with a constant value. 
We use warper $W$ to warp each garment into an approximate location on the original model image $m$. To produce $m^i$, we paste each warped garment onto $m$ and fill in the skin regions with the median skin tone of the face using the predicted parsing $p$ from the Semantic Layout Generator $H$ (see Fig.~\ref{fig:diffusion_inference}). 

As shown in Zhang~\etal~\cite{Zhang2024Preserving}, a good starting point can significantly improve generation quality. Hence, instead of starting the inference procedure from pure noise, we start with a noisy incomplete image by adding noise to $m^i$. Additionally, we use $m^i$ as the control and CLIP image embedding~\cite{ramesh2022} (see Fig.~\ref{fig:training_architecture}). Finally, we concatenate the smplx joint image $j$~\cite{pavlakos2019smplx} to the input of the denoiser to guide where the skin will be generated. The full body inference pipeline is shown in the top path of Fig.~\ref{fig:diffusion_inference}. 

Additionally, we can generate any zoomed close-up region of the try-on image by cropping a smaller region in our incomplete model image $m^i$ and its corresponding smplx joints image $j$. We then upsample the cropped region to the image resolution of $m^i$ to get $m^i_{zoom}$ and pass $m^i_{zoom}$ to our diffusion network $F$ to generate high-resolution close-ups of the try-on result. The zoom inference procedure is shown in the bottom path of Fig.~\ref{fig:diffusion_inference}.

\begin{figure}[tb]
  \centering
  \begin{subfigure}{0.49\linewidth}
    \includegraphics[width=1.0\textwidth]{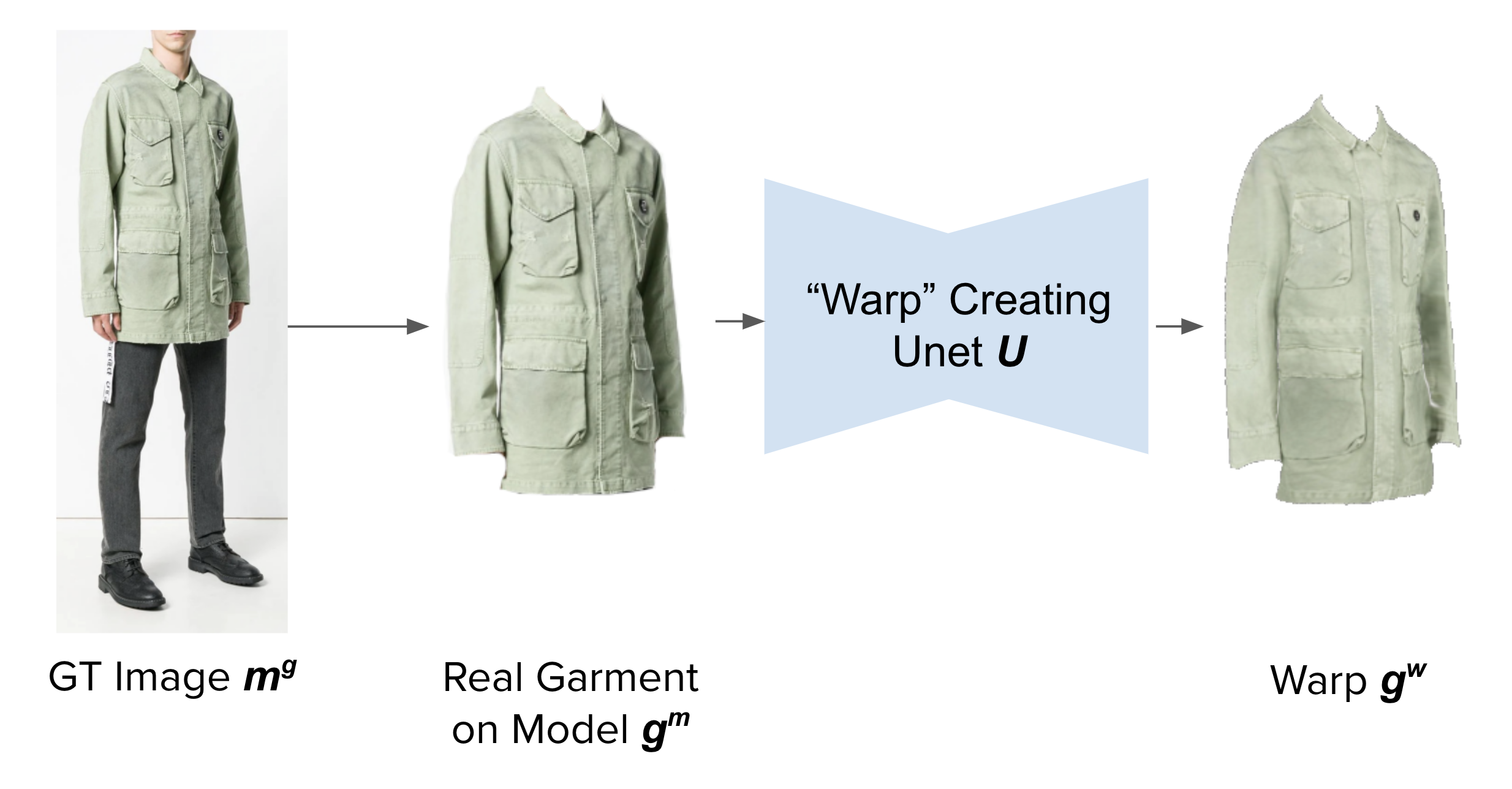}
    \caption{Training $U$ to Convert $g^m$ to Look Like $g^w$}
    \label{fig:warp_training(a)}
  \end{subfigure}
  \hfill
  \begin{subfigure}{0.49\linewidth}
    \includegraphics[width=1.0\textwidth]{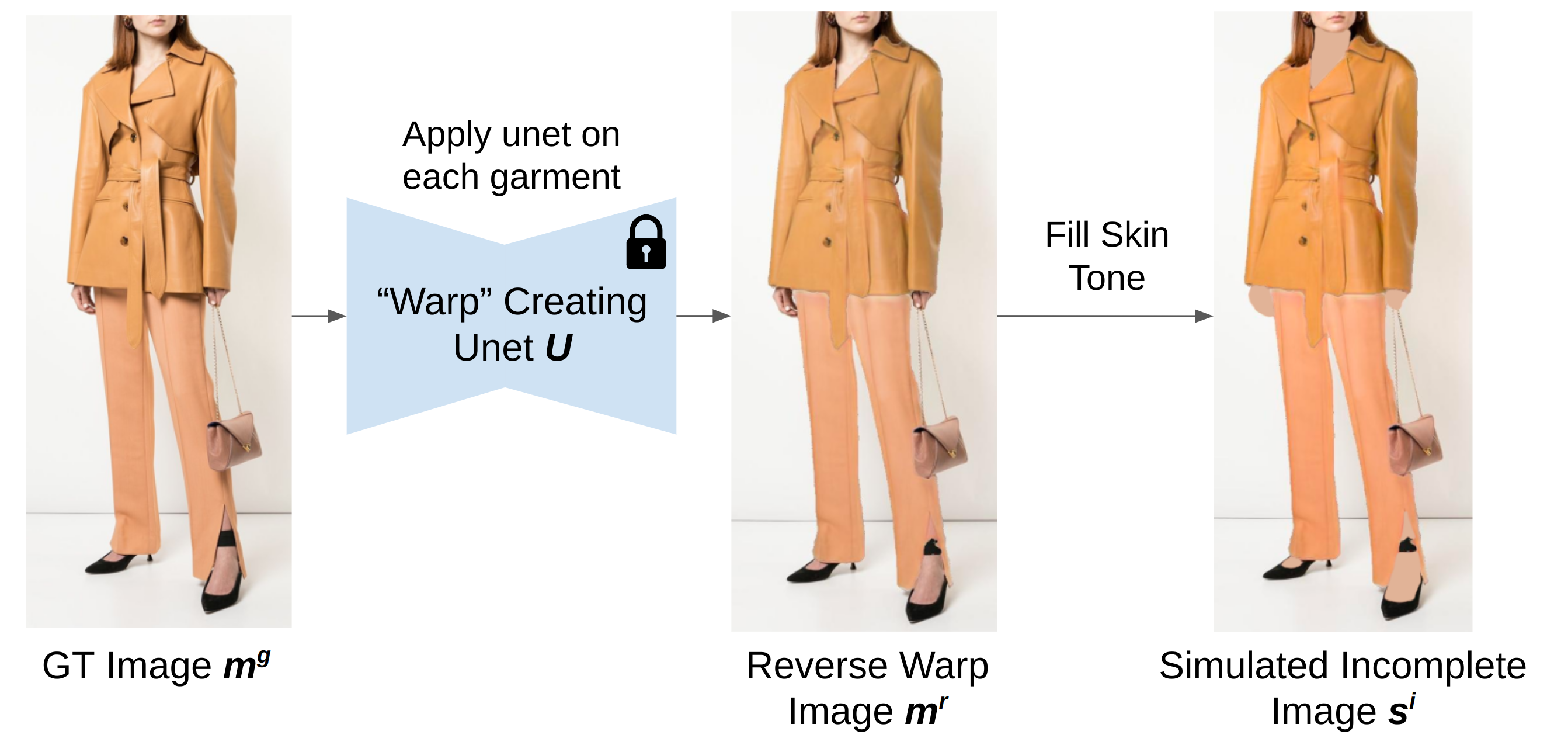}
    \caption{Creating Simulated Incomplete Image $s^i$}
    \label{fig:warp_training(b)}
  \end{subfigure}
  \caption{We show how the simulated incomplete image $s^i$ is created for training denoiser $F$. 
  The first step is to train a U-Net~\cite{UNet} to take a garment in a model image and convert it to a warp distribution. The next step is to run the U-Net on each garment in a person image to create "reverse warps" for each garment. This alters each garment to look like warped garments, but crucially, all the garment features are aligned. Finally, the incomplete image is created by filling in the skin with a constant skin value derived from the median pixel value of the face.}
  \label{fig:warp_training}
\end{figure}

\subsection{Training Procedure}
\label{sec:training_method}
The key to obtaining good accuracy is ensuring the diffusion model does not hallucinate unnecessary details.  This requires a careful training procedure because the garment features in $m^i$ produced from warps tend not to be precisely aligned with the garment features in the ground truth image.  A diffusion model trained with $m^i$ interprets small misalignments as encouragement to hallucinate. To avoid this effect, we train the diffusion model with simulated $m^i$ in which the garment features are precisely aligned. 

\subsubsection{Simulated $m^i$ for Containing the Diffusion Model.}
\label{sec:incomplete_image}
During training, we create a simulated incomplete image $s^i$ by removing detail from the ground truth model image $m^g$ rather than using garment warps. This allows the simulated incomplete image's high-frequency features, like text, patterns, textures, etc., to perfectly align with the ground truth image (Sec.~\ref{sec:ablations} highlights the importance). 

The first step is to train a "warp" creating network $U$ to take a real garment on a model and convert it to the warp distribution. This is done by feeding in garments segmented off a model $g^m$ and training $U$ to convert it to a warped garment $g^w$ created from a pre-trained warper $W$ (see Fig.~\ref{fig:warp_training(a)}). Notice in Fig.~\ref{fig:warp_training(a)} that the garment image $g^m$ differs from that of the warp $g^w$ (less shadows, wrinkles, etc.), demonstrating a difference between the two image distributions. We apply L1 loss and adversarial loss to train this network. 

The second step is to run our trained $U$ on each garment in a model image to create a reverse warp model image $m^r$. This simulates the look of a warped garment by removing details like shadows, wrinkles, and folds. Crucially, the essential features of the garments (e.g. buttons, pockets, collars, straps, logos, etc.) in $m^r$ are still aligned with the garments in $m^g$ (see Fig.~\ref{fig:warp_training(b)}). Finally, the simulated incomplete image $s^i$ is created by filling in the skin of the model in $m^r$ with a constant skin value derived from the median pixel value of the face (see Fig.~\ref{fig:warp_training(b)}). We now have training pairs of ground truth model images $m^g$ and aligned simulated incomplete images $s^i$, where the garments in $s^i$ are similar in distribution to warped garments with details such as skin, shading, textures, etc. removed. 

\subsubsection{Training the Diffusion Network.}
\label{sec:incomplete_model_training}
We train our diffusion model $F$ with the simulated incomplete image $s^i$ as a control to recover the ground truth image $m^g$. The diffusion model learns to make $s^i$ more realistic by adding texture details, shadows, skin, and other modifications. We call this training \textbf{Simulated Incomplete Image Training}. 

We adopt a ControlNet-based~\cite{Zhang2023ControlNet} architecture, feeding in the control image $s^i$ as control and initializing the weights with pre-trained Stable Diffusion~\cite{rombach2021highresolution} (see Fig.~\ref{fig:training_architecture}). Rather than using text embeddings as the condition, we use the CLIP Image encoder~\cite{ramesh2022} to take $s^i$ and convert it to an embedding to feed into intermediate layers of the denoiser through cross-attention. Intermediate features from the control encoder are also combined with the denoiser's decoder with cross-attention. Finally, we concatenate smplx joints image $j$ to the input of the denoiser to help guide the network generate skin.

In Zhang~\etal~\cite{Zhang2024Preserving}, the authors revealed that the difference in training initialization (ground truth image + noise) and inference (pure Gaussian noise) led to unexpected artifacts in inference generations. To make the training and inference initialization consistent, we adapt their procedure by using our control image $s^i$ as initialization. We call this modified procedure \textbf{Control Initialization}, where we use our noisy control image as the initialization during training and inference. This training modification is applied to the first $S$ time steps (details on the loss function are in the Supplementary).
\begin{figure}[tb]
  \centering
    \includegraphics[width=1.0\textwidth, trim=4 4 4 4,clip]{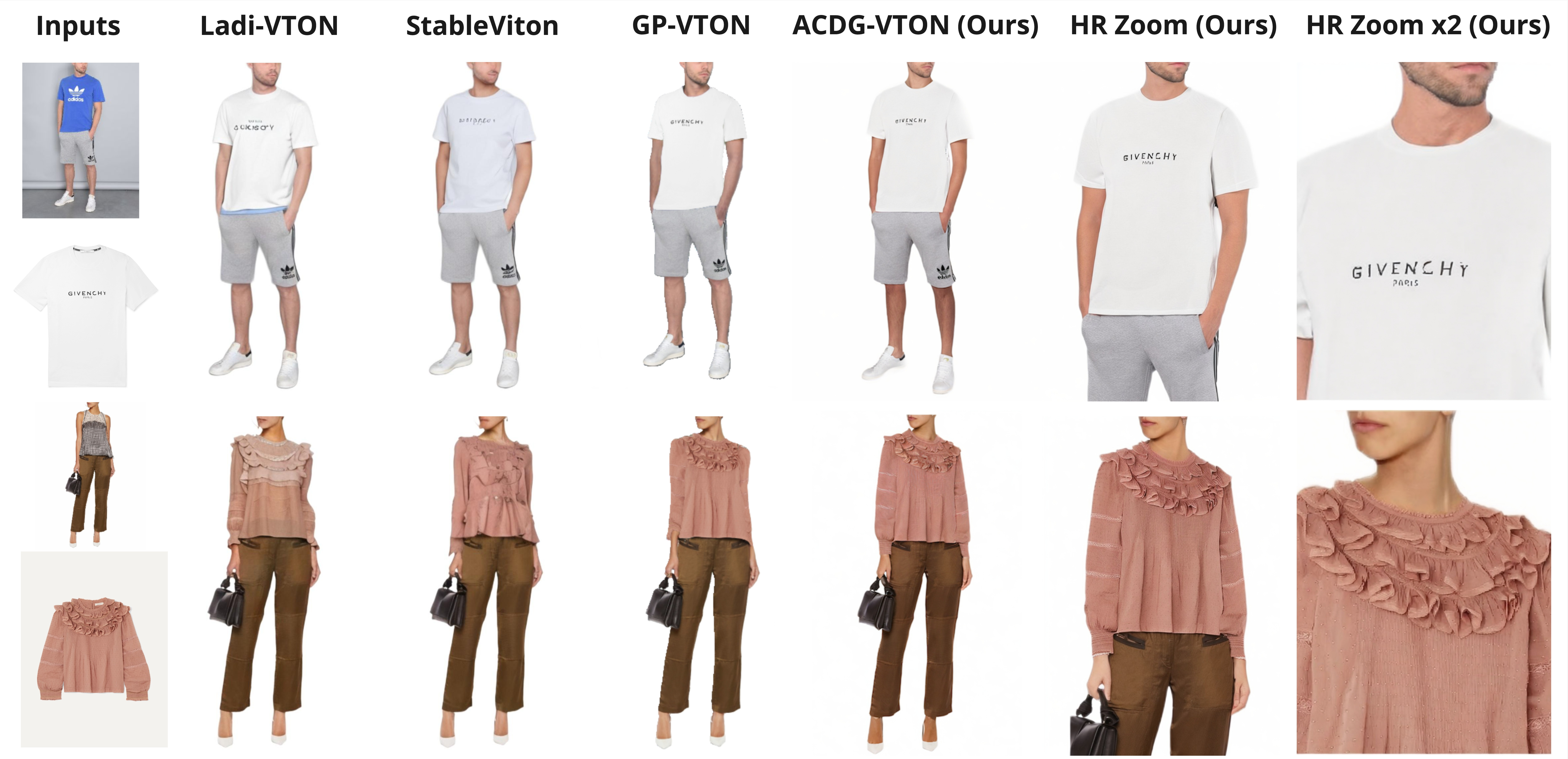}
  \caption{We show more qualitative examples on DressCode and include GAN-based warping method GP-VTON~\cite{Xie2023GPVTON} for comparison. Our method preserves garment details and textures more accurately. While our method faithfully copies details, certain high-frequency details are still modified due to using a VAE. For instance, the "Paris" under "Givenchy" is distorted. However, if we run our HR Zoom method and generate a high-res cropped region, we can recover the spatial details of the text, something other methods cannot do. Similarly, fine-grained details (i.e. the dots) on the bottom row pink blouse only appear in our zoom results.}
  \label{fig:sota_comparison}
\end{figure}

\subsubsection{Training High-Resolution Zoom.}
\label{sec:zoom_training}
Our method enables us to produce high-resolution zoom (\textbf{HR Zoom}) in a try-on image without the need to train at higher resolutions. This is possible because we ensure the alignment of features in the control image $s^i$ and the ground truth model image $m^g$ during training. During inference, when we crop and upsample regions of $m_i$ to get $m^{i}_{zoom}$, the denoiser $F$ will accurately copy details from the reference $m^{i}_{zoom}$ and improve the resolution. This process can help overcome high-frequency distortions by the VAE by moving high-frequency details into lower frequencies, as depicted in Fig.~\ref{fig:vae_encode_decode}. To support this feature, we randomly crop and upsample regions of $s^i$ and its corresponding features in the training process.

\section{Datasets and Training}

\subsection{Datasets}
We train our network on the OVNet dataset~\cite{Kedan_Li_2021_CVPR}, which contains garments of different categories worn in different styles. For fair comparisons against other methods, we also evaluate on DressCode~\cite{morelli2022dresscode}. Due to commercial licensing restrictions, we expressly avoid training our model on any images from DressCode; we solely use DressCode images for testing and comparison purposes and not for any commercial use. Since our model is trained on OVNet's dataset and tested on DressCode, differences between the two datasets may make the testing process more challenging for our method. Despite this, we evaluate the upper-body case in DressCode to account for the upper-body constraints of other methods.

\subsection{Training Details}
We take a Semantic Layout Generator $H$ and the Warper $W$ from Rendering Policies~\cite{Li2024Controlling}, but any pre-trained warper and layout generator from other methods can work (e.g.,~\cite{tprvton,Issenhuth2020DoNM, Kedan_Li_2021_CVPR, Chopra_2021_ICCV}). We use a resnet18-based unet for our "warp" creating unet $U$ and train 1 epoch over the OVNet dataset. We use batch size 32, learning rate 1e-4, and Adam optimizer~\cite{adam}. For denoiser $F$, we initialize our weights from Stable Diffusion v1.5~\cite{rombach2021highresolution} and make a copy of the encoder weights for our control network. During training, we freeze the encoder of the denoising U-Net to preserve learned features of the pre-trained model and unfreeze all other weights. We train on image size 1024x768 with batch size 4 and use learning rate 1e-5 with Adam optimizer for 1 epoch on the OVNet dataset. We set $S=50$, and all inference is done with 20 diffusion steps. To support zoom-in generations, we randomly crop and add padding to $m^i$ and its corresponding smplx image $j$ between a scale of 0.25 and 2.0.


\section{Experimental Evaluation}
\label{sec:evaluation}
\begin{figure}[tb]
  \centering
    \includegraphics[width=0.92\textwidth]{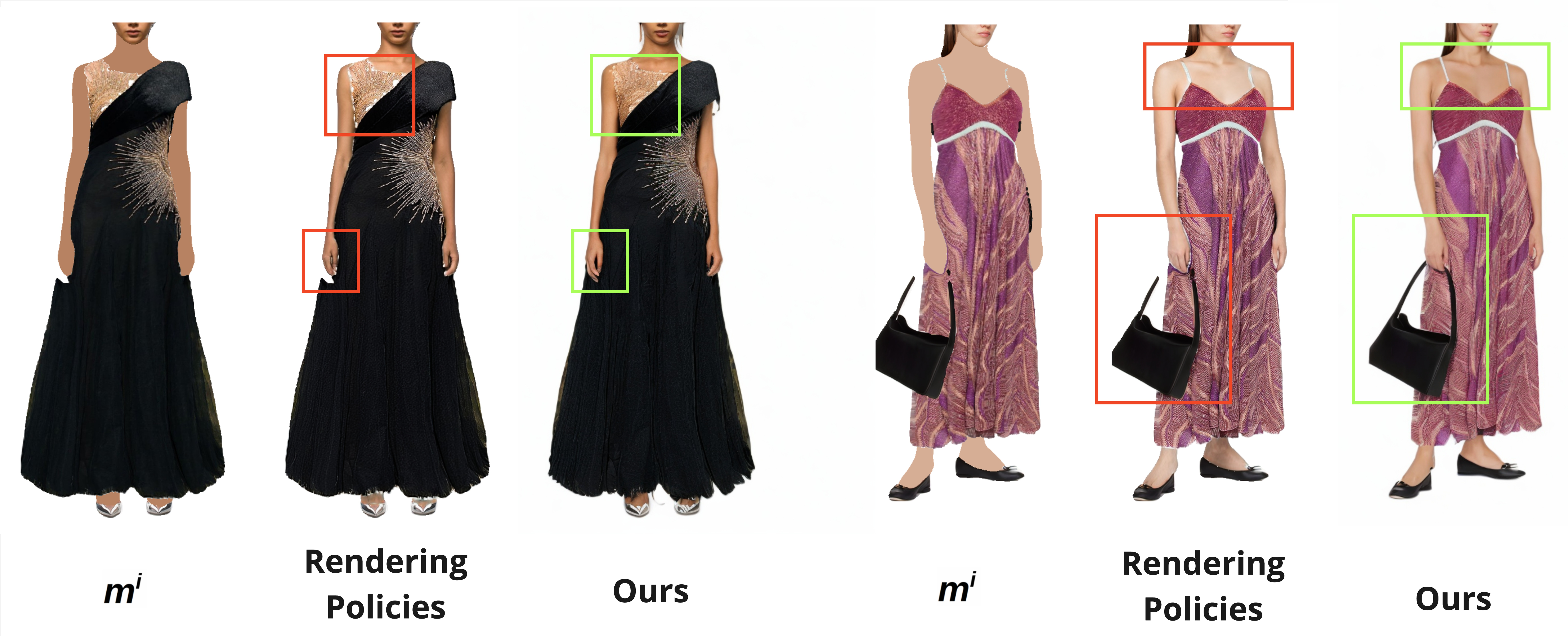}
  \caption{Our method produces higher quality images (where quality refers to the generated images' realism and visual appeal). 
  Note improvements over Rendering Policies in small artifacts, minor warp errors, and segmentation issues (red box: issue; green box: improved). Figures show $m^i$ in each case for reference.}
  \label{fig:rendering_comparison}
\end{figure}

We engineered our system to keep the advantages characteristic to Rendering Policies~\cite{Li2024Controlling} such as good accuracy and garment controllability, which we showcase through qualitative demonstrations. The limits of current image generation metrics necessitate user studies and qualitative assessments, but we still provide standard metrics for reference (Table~\ref{tab:metrics}). We show our method effectively preserves garment details with no significant difference in accuracy compared to~\cite{Li2024Controlling} (as exhibited in Fig.~\ref{fig:rendering_comparison} and Supplementary). However, we show our system archives much better quality than~\cite{Li2024Controlling} through qualitative examples and a user study (Sec.~\ref{sec:results}). More importantly, our method outperforms other state-of-the-art baselines significantly in terms of accuracy, shown through two user studies designed to focus on accuracy (Sec.~\ref{sec:results}). We evaluate not only full-body VTON images but also challenging garment close-up images for VTON, showing our method preserves accuracy better than other baselines. 

\subsection{Evaluation Procedures}
\label{sec:evaluation_procedures}
\subsubsection{SOTA Comparisons.} In our evaluations, we compare several types of state-of-the-art works: LaDI-VTON~\cite{Morelli2023LadiVTON} (warp-based diffusion), StableVITON~\cite{Kim2023StableVITON} (implicit warp-based diffusion), and GP-VTON~\cite{Xie2023GPVTON} (warped-based GAN). We take all method's released models from their official GitHub repositories. Similar to our evaluation scenario, StableVITON's weights are trained on another dataset instead of DressCode, making their evaluation more challenging. However, the paper emphasizes their weights are generalizable across different datasets in their cross-dataset analysis. We run standard metrics LPIPS~\cite{zhang2018perceptual} and SSIM~\cite{SSIM} for the paired setting, and FID~\cite{NIPS2017_8a1d6947} and KID~\cite{Binkowski2018DemystifyingMG} for the unpaired setting. Because GP-VTON and our method generate white backgrounds, we remove the background for all generations when running the metrics for fair comparison. Finally, we compare to Rendering Policies~\cite{Li2024Controlling} because we engineered our system around its warper and layout generator.

\subsubsection{User Studies.} We designed three user studies to capture differences between accuracy and quality. We designed two accuracy user studies to compare the accuracy between \cite{Morelli2023LadiVTON, Kim2023StableVITON, Xie2023GPVTON} and ours on DressCode~\cite{morelli2022dresscode} in two scenarios: regular full-body VTON generation and our novel zoomed-in VTON generation. We asked participants to "Choose the option that most accurately shows the model wearing the new garment (see arrows for guidance)." Because differences can be hard to spot, we added arrows to regions on the garment that may be hard to notice so users can focus on the accuracy of the features (see Supplementary for example). For full-body generations, we show all results in 512x384 resolution as most baselines only released 512x384 resolution models. We ran an additional user study to compare our method's ability to zoom into regions. For our method, we zoomed into the top half of the model and showed results in 367x275. For other works, we cropped the same upper 367x275 region out of their 512x384 results. Finally, for our quality user study, we compared 1024x768 results between Rendering Policies~\cite{Li2024Controlling} and Ours on full outfit data from OVNet~\cite{Kedan_Li_2021_CVPR}. We asked participants, "Here are two computer-generated images with slight differences. Which image do you think is better overall?". We evaluate against~\cite{Li2024Controlling} for quality, but not for accuracy because we expect any difference to be small (see Fig.~\ref{fig:rendering_comparison} and Supplementary). More details for all user studies are located in the Supplementary.

Our primary interest is in the choice of method. For each comparison example, we identify the method most users prefer and count the number of examples where that method was preferred. Note that this means that a method preferred by some users on some examples may still score 0 (as StableVITON does, Fig.~\ref{fig:user_study}).

\begin{figure}[tb]
  \centering
    \includegraphics[width=0.98\textwidth]{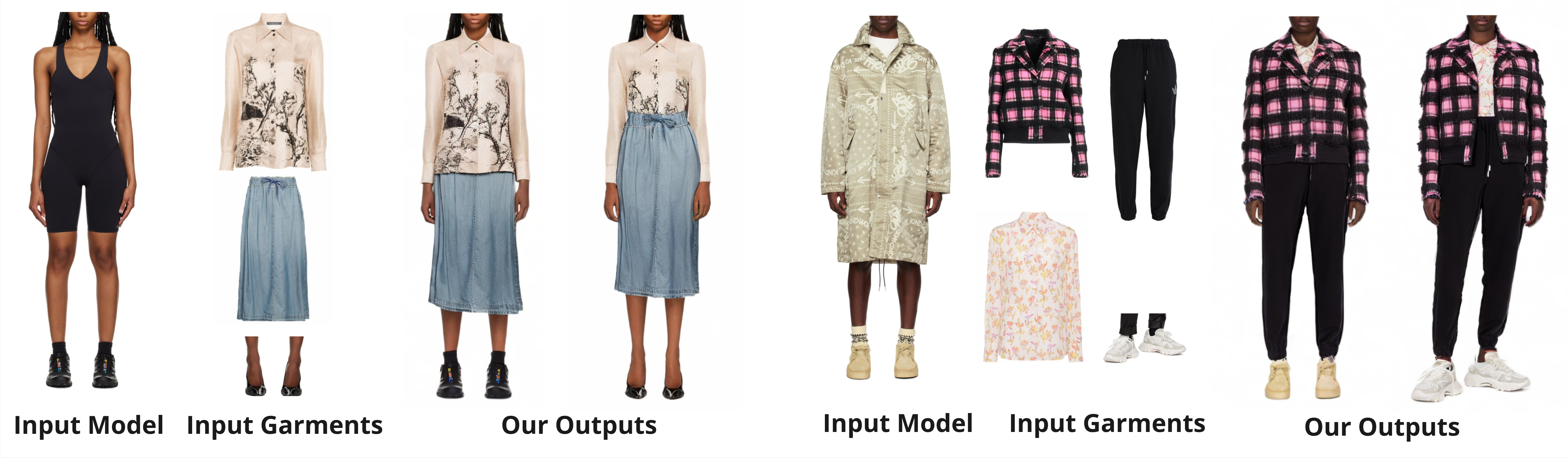}
  \caption{Garment controllability refers to the ability to control how the garment is styled. Because our method respects the warp, we can inherit any warp-based method that has controllability of garments (e.g. Rendering Policies~\cite{Li2024Controlling} or Size Does Matter~\cite{Chen2023Size}). Our method can layer multiple garments, apply styling options such as tuck/untuck and open/close outerwear, and even support different shoes to complete the look. }
  \label{fig:use_cases}
\end{figure}

\subsection{Results}
\label{sec:results}
\textbf{Qualitatively, our method preserves garment details}, accurately generating details such as texts, logos, and patterns. In Fig.~\ref{fig:teaser}, we compare our method against state-of-the-art diffusion methods on DressCode. We show our method replicates the leopard pattern in the first row and is able to display "Neighborhood 94" in the second row accurately. In contrast, diffusion-based methods like LaDI-VTON and StableVITON alter and hallucinate details on the garments. Furthermore, our HR Zoom method can generate high-resolution cropped regions and preserve details the VAE may distort (see Fig.~\ref{fig:vae_encode_decode}). 

Our training method resolves two accuracy issues associated with diffusion methods: the hallucination of garment details and the elimination of high-frequency details. First, our method prevents hallucination of details because the simulated incomplete image $s^i$ aligns with the ground truth image $m^g$ during training.
As a result, the diffusion network learns to faithfully copy the garment features of the warped items during inference. Second, our method addresses the high-frequency detail removal from VAEs by cropping and upsampling specific regions of $m^i$ before feeding into our denoiser $F$. We successfully overcome VAE limitations without training a model in higher resolution.
In Fig.~\ref{fig:sota_comparison}, we visualize additional comparisons on DressCode. We show our extremely zoomed-in generations can preserve even the smallest details on the garments (i.e. the "Paris" text under "Givenchy" on the white t-shirt and even the dot textures in the pink blouse), something other methods cannot accomplish. 

\begin{figure}[tb]
  \centering
  \begin{subfigure}{0.45\linewidth}
    \includegraphics[width=1.0\textwidth]{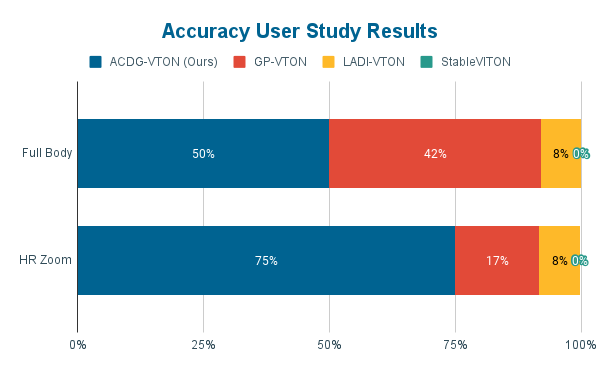}
    \caption{Accuracy User Study Results}
    \label{fig:user_study(a)}
  \end{subfigure}
  \hfill
  \begin{subfigure}{0.45\linewidth}
    \includegraphics[width=1.0\textwidth]{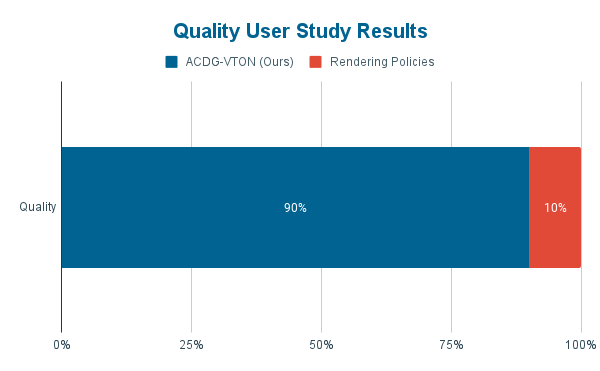}
    \caption{Quality User Study Results}
    \label{fig:user_study(b)}
  \end{subfigure}
  \caption{Our method outperforms baselines in accuracy and quality in user studies with ~40 participants each. For accuracy, we compare ours to LaDI-VTON~\cite{Morelli2023LadiVTON}, StableVITON~\cite{Kim2023StableVITON}, and GP-VTON~\cite{Xie2023GPVTON} and outperform these baselines in both the normal full body scenario and our novel zoom-in scenario. For quality, we compare ours to Rendering Policies~\cite{Li2024Controlling} and see a substantially larger preference in quality.}
  \label{fig:user_study}
\end{figure}

{\bf Quantitatively, our method preserves garment details}, outperforming strong competitors in user studies. In Fig.~\ref{fig:user_study(a)}, we see both in the full body and the zoom-in case, users believe our method better preserves details compared to baselines. While GP-VTON is close in full body performance, when we zoom into details, users can easily notice details that GP-VTON misses and our method better preserves. It verifies that our method copies details faithfully even when zoomed in and re-generating at the zoom-in resolution helps alleviate VAE encoding alterations. Finally, in Table.~\ref{tab:metrics} we show our method outperforms all other baselines on metrics such as Lpips, SSIM, FID, and KID.

Although we used the published version of code and weights from~\cite{StableVITONGithub}, StableVITON underperforms. We see two possible reasons.  First, StableVITON's weights were trained on a different dataset instead of DressCode.  Our method was trained this way, too, but StableVITON may have more trouble than our method with this generalization.  Second,  the demanding "winner-takes-all" form of the user study advantages a system that does well on most examples.


\begin{table}[tb]
\centering
\caption{We compare standard metrics against different methods on DressCode~\cite{morelli2022dresscode} upper body with white background. We outperform prior work across all metrics.}
\begin{tabular}{ c c c c c }
\hline
\textbf{Method} & \textbf{LPIPS($\downarrow$)} & \textbf{SSIM($\uparrow$)} & \textbf{FID($\downarrow$)} & \textbf{KID($\downarrow$)}\\
\hline
 GP-VTON~\cite{Xie2023GPVTON} &  0.2154 & 0.7953 & 14.754 & 5.485\\
 LaDI-VTON~\cite{Morelli2023LadiVTON} &  0.0671 & 0.9260 & 16.098 & 3.509\\
 StableVITON~\cite{Kim2023StableVITON} &  0.0654 & 0.9271 & 17.500 & 5.070\\
\hline
 \textbf{ACDG-VTON (Ours)} &  \textbf{0.0579} & \textbf{0.9281} & \textbf{14.750} & \textbf{1.940}\\
 \hline
\end{tabular}
\label{tab:metrics}
\end{table}

\textbf{Our method enhances quality} compared to GAN-based methods while maintaining the accuracy of garments. As depicted in Fig.~\ref{fig:rendering_comparison}, GAN-based methods like Rendering Policies struggle to rectify issues with minor artifacts, warping problems, and poor segmentations. On the other hand, our method can institute realistic modifications without tampering with the essential details of the garments present in the warped inputs. This is attributed to the pre-trained diffusion process’s ability to decipher and implement elements of realism, enabling us to use diffusion to correct minor issues in the input that it identifies as unrealistic. In Fig.~\ref{fig:user_study(b)}, user studies indicate that participants overwhelmingly thought the quality of our method exceeded that of GAN-based Rendering Policies. 
Artifacts are minimized because they can be fixed by diffusion, whereas GANs are less flexible in fixing errors it does not see regularly during training (see Fig.~\ref{fig:rendering_comparison}).

\textbf{Our method has versatile garment controllability}, allowing users to style different garments in an outfit.  In Fig.~\ref{fig:use_cases}, we show our method allows control over styling options such as tuck-in/tuck-out tops and open/closed outerwear. In addition, our method allows layering any number of garments and allows for try-on with different shoes.

This controllability results from how we contained diffusion to respect the warped garments in the control image $m^i$ during inference. By adjusting the order and size of the garments with explicit warps, we control the generation through the control image. The control image acts as a canvas to dress the model with any number of garments, alter the style, and even add shoes, while only requiring a single diffusion inference cycle (20 timesteps) to generate the final image.

\begin{figure}[tb]
  \centering
    \includegraphics[width=1.0\textwidth]{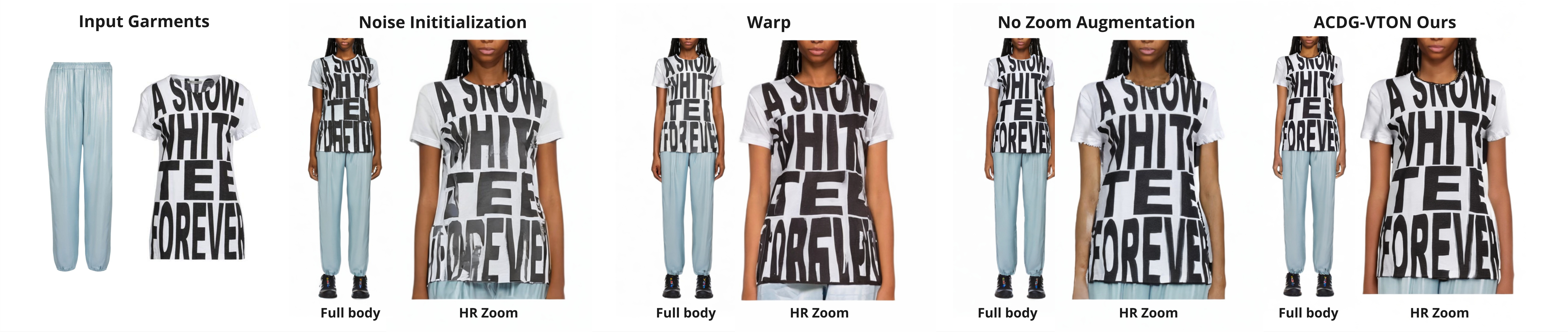}
  \caption{Each component of our system (Sec.~\ref{sec:incomplete_model_training}) is important. Standard training and inference (i.e. omitting Control Initialization) leads to very poor text ("Noise Initialization"). Using warped garments rather than simulated incomplete images $s^i$ during training leads to texture problems ("Warp"). No zoom augmententation in training produces blurry text boundaries and artifacts on arms, neck, and sleeves for HR Zoom examples ("No zoom augmentation").
  }
  \label{fig:ablations}
\end{figure}

\subsection{Ablations}
\label{sec:ablations}

\textbf{With vs Without Control Initialization.}
In Fig.~\ref{fig:ablations}, we compare Control Initialization against standard diffusion training and inference initialization with noise~\cite{song2020denoising} ("Noise Initialization" figure). The results of noise initialization are considerably less stable and the generations for both full-body and HR zoom show significant deformations to the text. 

\textbf{Simulated Incomplete Image vs Warp-based Image Training.}
Instead of using $s^i$ as the control during training, we show results of using the incomplete model image $m^i$ as the control where the details of the garments are not perfectly aligned (created by garment warps). Notice in the "Warp" figure in Fig.~\ref{fig:ablations} that there are alterations to the text in both the full-body and zoom generations. This highlights in the warp-based training case, diffusion learns to hallucinate details rather than accurately copy the reference.

\textbf{With vs Without Zoom Augmentation.}
In the "No Zoom Augmentation" figure in Fig.~\ref{fig:ablations}, we show results without zoom augmentation during training. The full-body generation looks good, as expected. However, without the zoom augmentation training, the generation for HR Zoom result has blurry text boundaries and artifacts on the arms, the neck, and the sleeves.

\section{Conclusion}
\label{sec:future}
We present a versatile diffusion method that checks the boxes for accuracy, quality, and garment controllability. We reveal issues with standard diffusion training, including limitations with using VAEs, that our proposed method can fix by aligning garment features perfectly during training and zooming in when needed. We show our diffusion model can fix artifacts in inputs and improve the overall quality of the image. Finally, we show a controllable pipeline that allows for styling, layering multiple garments, and try-on for shoes.


%
%
\bibliographystyle{splncs04}
\bibliography{egbib}

\clearpage
\begin{center}
\textbf{\Large ACDG-VTON: Accurate and Contained Diffusion
Generation for Virtual Try-On}\\
\textbf{\large Supplementary}
\end{center}
\setcounter{equation}{0}
\setcounter{figure}{0}
\setcounter{table}{0}
\setcounter{section}{0}
\makeatletter
\renewcommand{\theequation}{S\arabic{equation}}
\renewcommand{\thefigure}{S\arabic{figure}}

\section{Control Initialization Implementation}
\subsection{Stable Diffusion Background}
Stable Diffusion~\cite{rombach2021highresolution} is trained to recover an image $x_0$ by denoising a noisy image $x_t$ at timestep $t \in [0,T]$. However, for computation efficiency, most diffusion-based methods are trained in the latent space of a VAE~\cite{Kingma2014}. Let $E$ represent the VAE encoder and $D$ represent the VAE decoder. Then, Stable Diffusion is trained to recover $z_0 = E(x_0)$ by denoising a noisy latent $z_t$ at time step $t \in [0, T]$. At the $t$'th timestep, the denoiser is presented with
\begin{equation} \label{eq:x_t}
    z_t = \sqrt{\alpha_t} z_0 + \sqrt{(1-\alpha_t)}\epsilon_t
\end{equation}
where $\epsilon_t \sim \mathcal{N}(0,I)$ and $\alpha_t$ is the cumulative product of scaling at each timestep $t$ (refer to DDIM~\cite{song2020denoising}). Following the training procedure from~\cite{song2020denoising}, a denoiser $F$ predicts the added noise $\hat{\epsilon}_t = F(z_t, t, e; \Theta)$, where $F$ is parameterized by $\Theta$ and takes in noisy latent $z_t$, timestep $t$, and conditional encoding $e$. We write $\hat{z}_{0}$ for the predicted ground truth latent derived from removing the predicted noise $\hat{\epsilon}_t$ from $x_t$. Our loss is  
\begin{equation} \label{eq:loss}
    \mathcal{L} = \mathbb{E}[||\epsilon_t - \hat{\epsilon}_t||_2^2]
\end{equation}

From Eq.~\ref{eq:x_t}, if we have the predicted $\hat{\epsilon}_t$ and the noisy latent $z_t$, we can derive the predicted ground truth latent $\hat{z}_0$ with
\begin{equation} \label{eq:x_0}
    \hat{z}_0 = (z_t - \sqrt{(1-\alpha_t)}\epsilon_t)/ \sqrt{\alpha_t} 
\end{equation}
The predicted latent $\hat{z}_0$ can be mapped back to the image space to get our predicted image by feeding the latent through the decoder $\hat{x}_0 = D(\hat{z}_0)$.

\subsection{Control Initialization Training Loss}
In Zhang~\etal~\cite{Zhang2024Preserving}, the authors revealed that the difference in training initialization (ground truth image + noise) and inference (pure gaussian noise) led to unexpected artifacts in inference outputs. To make the training and inference initialization consistent, we adapt their modified training procedure by using noisy control images, $s^i$ and $m^i$, as the initialization during training and inference, respectively. Let $S$ be the number of skips per timestep during inference, $z^i = E(s^i)$, and $z_0 = E(m^g)$. As done in \cite{Zhang2024Preserving}, we apply the change to timesteps within the first skip step to guarantee the first skip step is trained with the noisy control image. Hence, we alter Eq.~\ref{eq:x_t} for $t \ge T - S$ to 
\begin{equation} \label{eq:x_t_new}
    z_{t}^{new} = \sqrt{\alpha_t} z^i + \sqrt{1-\alpha_t}\epsilon_t
\end{equation}
and 
\begin{equation} \label{eq:epsilon_new}
    \epsilon_{t}^{new} = (z_{t}^{new} - z_0 \sqrt{\alpha_t})/ (\sqrt{1-\alpha_t})
\end{equation}
Thus, our final loss is a combination of Eq.~\ref{eq:loss} and the new noise objective from Eq.~\ref{eq:epsilon_new}:
\begin{equation} \label{eq:new_loss}
\mathcal{L}_{new} = 
\begin{cases}
        \mathbb{E}[||\epsilon_t - \hat{\epsilon}_t||_2^2] & \text{if } t < T - S\\
        \mathbb{E}[||\epsilon_{t}^{new} - \hat{\epsilon}_t||_2^2] & \text{if } t \ge T-S
\end{cases}
\end{equation}

\section{Simulated Incomplete Image Examples}
\begin{figure}[tb]
  \centering
    \includegraphics[width=1.0\textwidth]{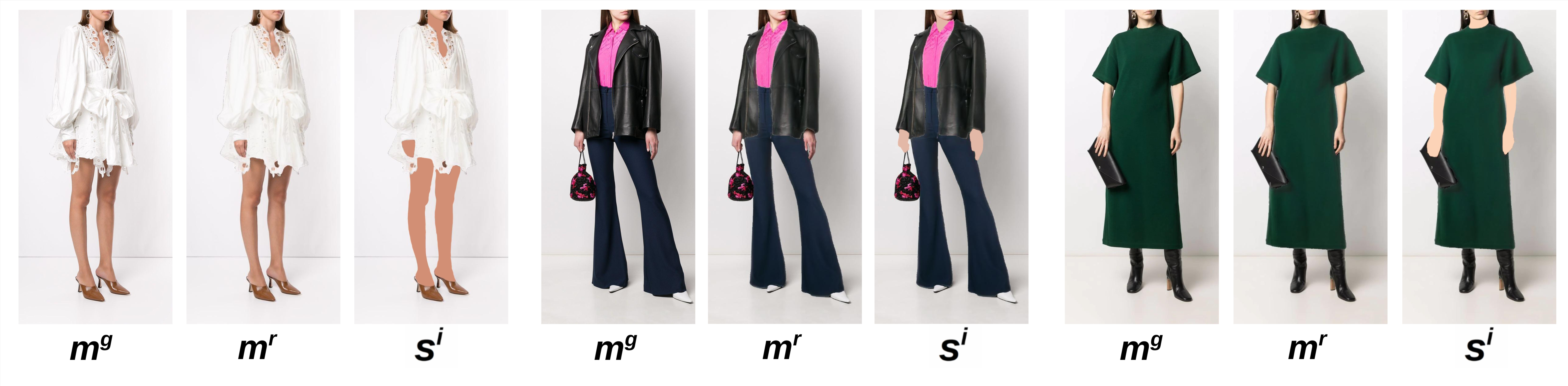}
  \caption{More examples of how reverse warp images $m^r$ and simulated incomplete images $s^r$ are created. Notice how the garments in $m^r$ have certain features altered around the edges of the garments. The other regions have shadows, folds, and wrinkles removed, but the features remain perfectly aligned.}
  \label{fig:simulated_image_supplementary}
\end{figure}

We provide detailed examples of the ground truth (GT) image $m^g$, the reverse warp image $m^r$, and the simulated incomplete image $s^i$ to illustrate that the essential features of the garments are aligned (Fig.~\ref{fig:simulated_image_supplementary}). Notice how the garments in $m^r$ have certain features altered around the edges to simulate how warped garments might look pasted on $m^g$. The other regions have shadows, folds, and wrinkles removed, but the features remain perfectly aligned.

\section{User Study}

\begin{figure}[tb]
  \centering
    \includegraphics[width=0.5\textwidth]{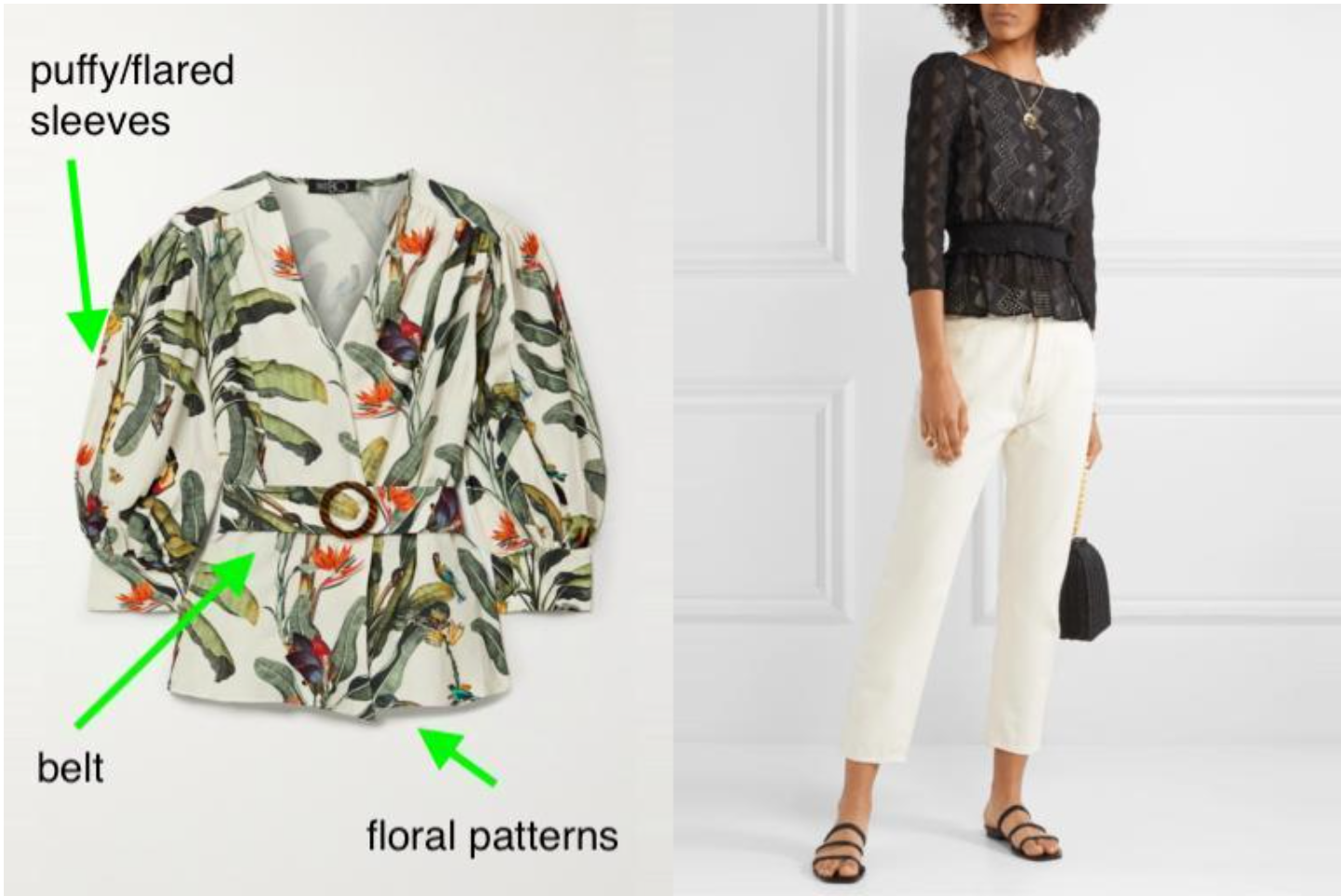}
  \caption{Users were prompted with arrows to help them focus specifically on the accuracy of the generations.}
  \label{fig:user_study_acc_image}
\end{figure}

For accuracy user studies, we compared the same random subset of 24 images from DressCode~\cite{morelli2022dresscode} upper body. To measure accuracy, we asked participants: "Choose the option that most accurately shows the model wearing the new garment (see arrows for guidance). If the options are similar, select the one that has the best overall quality (shadows, skin, realism, etc.)." We also provided an image of the model and garment with arrows pointing to garment attributes they may try to take note of (see Fig.~\ref{fig:user_study_acc_image}). We did this to encourage users to focus on the accuracy of garment attributes and to help those with an untrained eye to focus on subtle features that may have been altered in the generations. We collected 36 responses for the full-body accuracy study and 43 responses for the zoom accuracy study.

For quality, we took a random subset of 20 outfits from the OVNet dataset. To measure quality between Ours and \cite{Li2024Controlling}, we asked participants: "Each question has two computer-generated images with slight differences. Your task is to pick the image that you think has better quality. Quality refers to being more realistic, having fewer artifacts, having better lighting, and having cleaner details." We collected 41 responses for this quality study.

\section{More Qualitative Results}
\subsection{Demanding Inputs}
\begin{figure}[tb]
  \centering
    \includegraphics[width=1.0\textwidth]{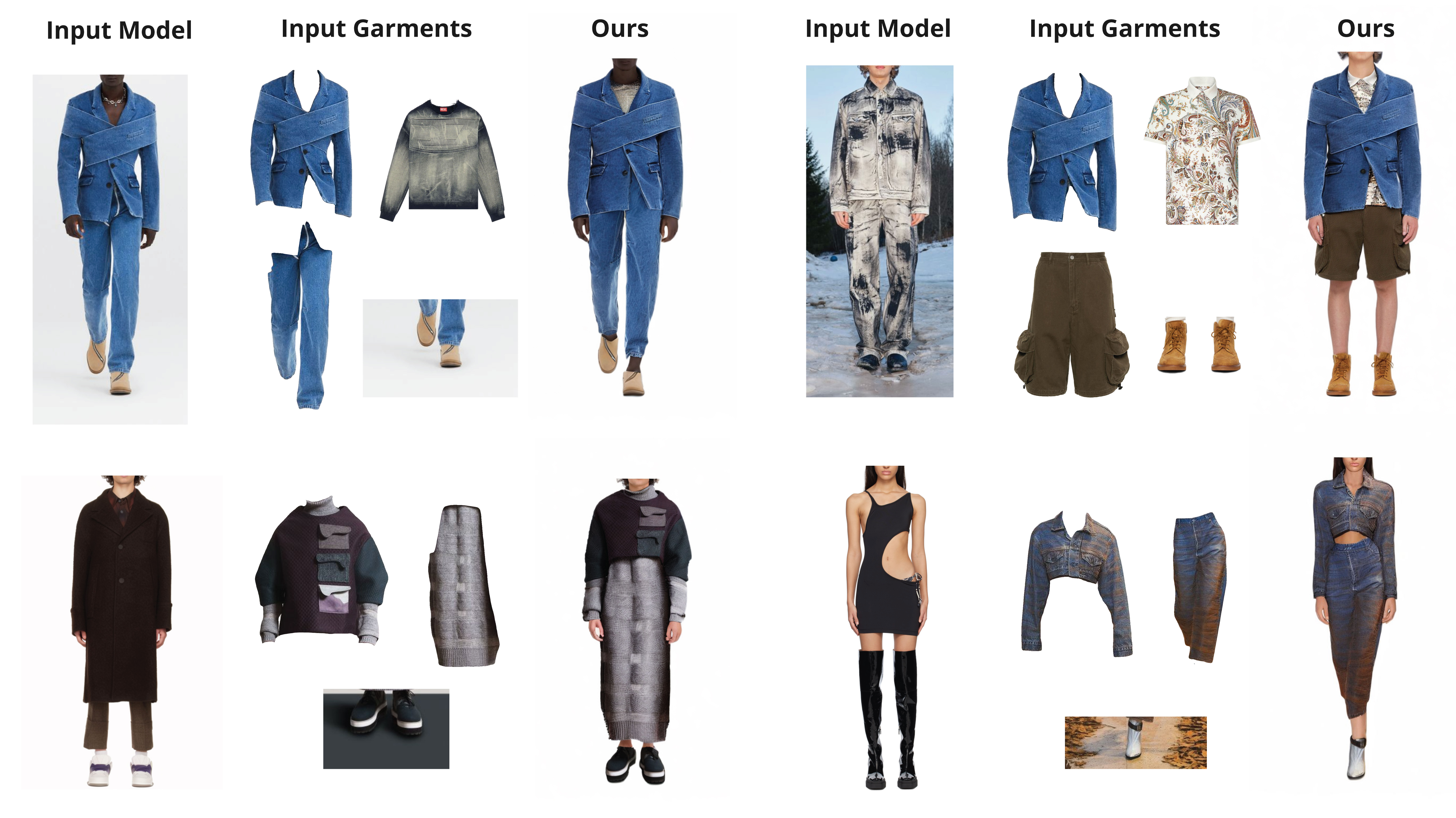}
  \caption{Qualitative examples demonstrate our method can operate on demanding inputs and user-segmented garments. Notice we can work with challenging user-segmented garments instead of studio-quality garment images.}
  \label{fig:use_cases_supplementary}
\end{figure}

In Fig.~\ref{fig:use_cases_supplementary}, we demonstrate that our method can work on challenging inputs like user-segmented garments, instead of standard studio garment images. In the top left example, we show we can digitize an outfit into our system by segmenting the top garment, bottom garment, and shoes from a person. If we want to add, for example, a shirt underneath the original outfit, we can generate the person with the segmented garments and the new shirt.  Note the garments in the generation do not look exactly as they do in the original input person because we are taking the segmented garments as input and feeding them through the warper $W$ and layout parser $H$. Hence, the entire result ("Ours") is predicted by our system based on the garment input images ("Input Garments"). This allows the user-segmented garments to be worn on any other model. In the top right example, we show that we can use the same segmented blue jacket and try-on the jacket on a different model with new garments and shoes.

In the bottom left example, we show a user-segmented top tucked into a user-segmented bottom. Notice that our method implements natural folds for the top to make it look like it is tucked into the bottom. Furthermore, despite the bottom having an irregular shape, our pipeline can infer details missing in the user-segmented garment to still produce a realistic version of the garment. Notice the pockets on the sweater and the stripes on the bottom are preserved.

Finally, in the bottom right example, we show our method can work with crop tops, asymmetric user-segmented pants, and even shoes that overlap. Notice that the cropped denim jacket is accurately portrayed as cropped outerwear and the person is realistically wearing the shoes, where one shoe is hidden behind the front shoe.

\begin{figure}[tb]
  \centering
    \includegraphics[width=1.0\textwidth]{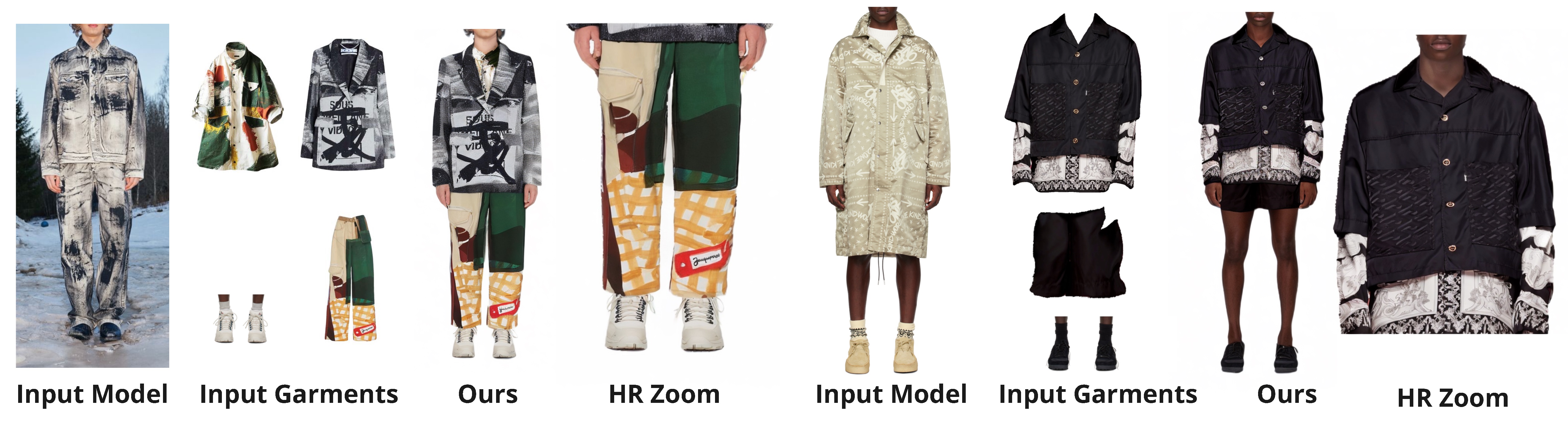}
  \caption{Our zooming procedure can correct high-frequency texture errors (from VAE spatial limitations), like the smudging of the text on the leg's label and the changes to the textures/patterns on the black and white shirt.}
  \label{fig:use_cases_zoom_supplementary}
\end{figure}

In Fig.~\ref{fig:use_cases_zoom_supplementary}, we show some more examples of our method operating with demanding inputs and user-segmented garments but focus on where HR-Zoom can correct issues with text and high-frequency textures due to the VAE's limited dictionary. Notice that our zooming procedure can correct texture features like the smudging of the text on the leg's label and the changes to the textures on the black and white shirt (see Fig. 2 in the main text for pointers on where the VAE alters details in the shirt). 

\begin{figure}[tb]
  \centering
    \includegraphics[width=1.0\textwidth]{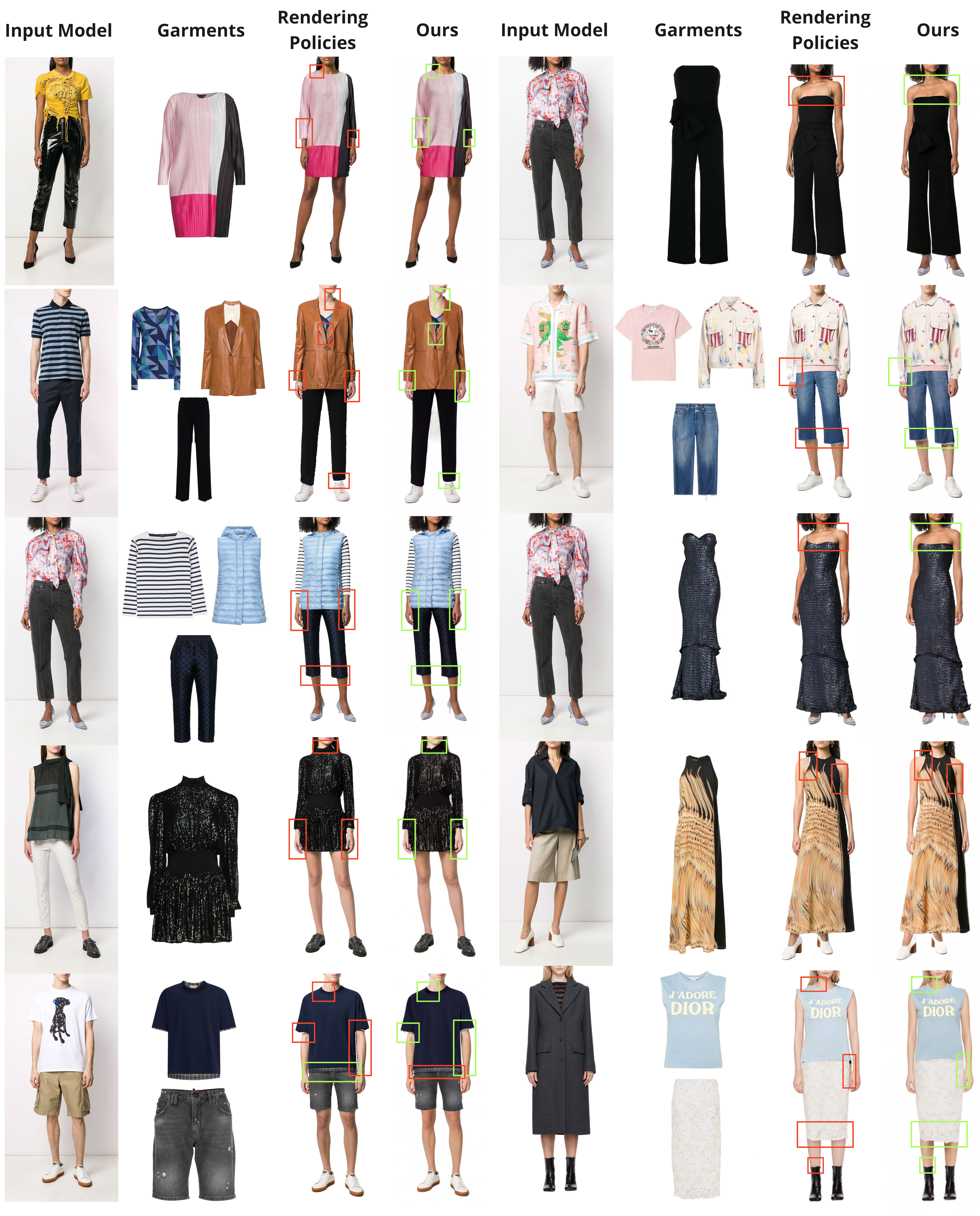}
  \caption{We compare our method against Rendering Policies for more examples in OVNet's dataset~\cite{Kedan_Li_2021_CVPR}. Notice accuracy is nearly identical, but our quality is better. Pay special attention to where the sleeves meet the arms, the necklines the tops, and the hems of the garments. Red boxes show issues, and green boxes show the issues fixed. Zoom in to see details.}
  \label{fig:rendering_comparison_supplementary}
\end{figure}

\subsection{Comparisons with Rendering Policies}
In Fig.~\ref{fig:rendering_comparison_supplementary}, we show our method is similar to Rendering Policies~\cite{Li2024Controlling} in accuracy but improves quality (we highlight some issues with red boxes and how they are fixed in green boxes). Pay close attention to how our method can fix bad connections between the legs/arms and the garments. Rendering Policies sometimes generates skin (hands, arms, legs, or neck) that is not connected to the garment due to bad layout predictions from the semantic layout parser $H$. Our method, ACDG-VTON, seamlessly rectifies these errors, making the image feel more natural and realistic.

In row 1 left, we see a small artifact near the neck and bad connections between the hands in Rendering Policies, whereas ACDG-VTON fixes these errors. 

In row 1 right and row 3 right, we see that Rendering Policies adds noticeable artifacts to the chest, possibly confusing the dress to have straps when there are no straps in the garment. In contrast, ACDG-VTON does not make this error and generates strapless dresses.

In row 2 left, we see bad connections between the hands and the foot, as well as a smudge at the chest where the outerwear and inner garment meet. ACDG-VTON generates a cleaner image without the smudge and fixes the connection between the hands/shoe and the garment to make the image look more realistic. Furthermore, notice that ACDG-VTON connects the outerwear naturally to the back of the neck and adds shadow underneath, making the garment look worn over the body rather than pasted on.

In row 2 right, we see a chunk of the person's right sleeve cut out in Rendering Policies, whereas ACDG-VTON fixes this hole. Furthermore, notice that Rendering Policies has a small bit of the person's calf protruding from the side of the jeans and the hem is not frayed. ACDG-VTON removes the protruding skin and adds a more realistic frayed look to the hem of the jeans.

In row 3 left, we see the Rendering Policies results' hands have small artifacts - the person's right hand has an unnatural white gap and the left hand has an unnatural gray gap. ACDG-VTON removes these gaps and fills them in naturally with the garment. Additionally, the connection between the calf and the pants has artifacts that ACDG-VTON seamlessly fixes. ACDG-VTON also adds natural soft shading to make the result feel more realistic.

In row 4 left, we see the same bad connections between the neck and the hands in Rendering Policies, where ACDG-VTON does not have the same errors. In addition, notice the gap between the hair and the dress at the neck in Rendering Policies that ACDG-VTON fills in with realistic detail. 

In row 4 right, we see more bad connections between the dress and the arms in Rendering Policies that ACDG-VTON seamlessly fixes.

In row 5 left, we see additional examples of bad connections between the shirt and the neck/arms in Rendering Policies, where ACDG-VTON shows better connections. We see that ACDG-VTON tries to add details to the sleeve's hem with the checked texture of the shirt to fill in the gap. While it is not completely accurate to the garment, it looks more realistic because the arms do not seem cut off. However, notice that ACDG-VTON does not represent the bottom hem of the shirt accurately compared to Rendering Policies. This is because the checkered pattern at the hem is a very fine texture pattern that ACDG-VTON cannot capture without our HR Zoom method. 

In row 5 right, we see several artifacts with skin protruding at the shoulder area, skin protruding at the hip, and a cut out of the person's right ankle in Rendering Policies. ACDG-VTON can fix all these artifacts and even makes the hem of the dress more accurate to the jagged hem cut of the skirt. 

\begin{figure}[tb]
  \centering
    \includegraphics[width=0.98\textwidth]{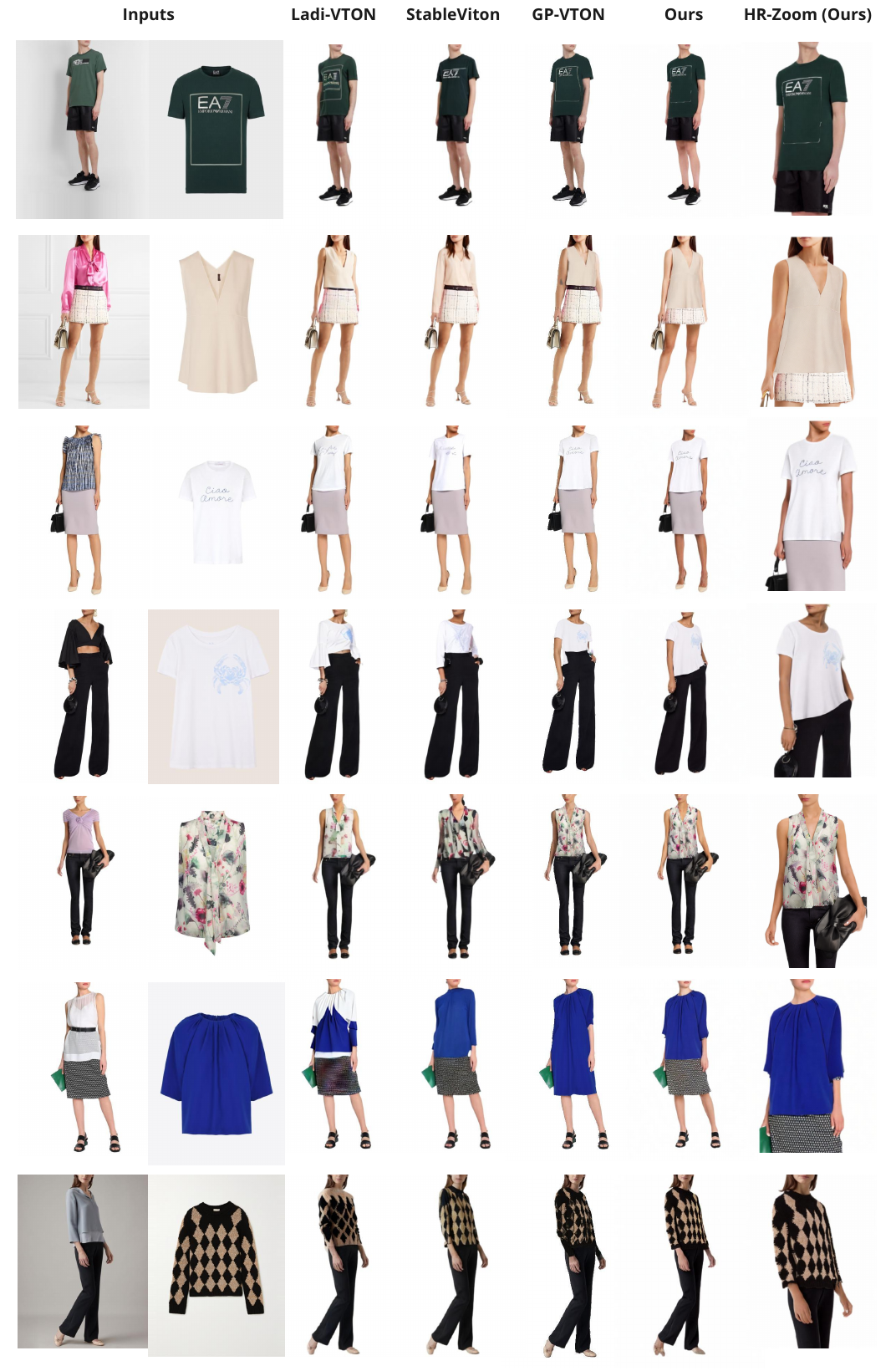}
  \caption{We show more Dresscode results compared to baselines at 512x384 resolution.}
  \label{fig:sota_comparison_supplementary}
\end{figure}

\subsection{More Comparisons with Baselines}
In Fig.~\ref{fig:sota_comparison_supplementary}, we show more comparisons against baselines~\cite{Morelli2023LadiVTON, StableVITONGithub, Xie2023GPVTON} on DressCode.  Results show our method can consistently generate the details more accurately compared to other SOTA methods.

In row 1, notice that Ladi-VTON and StableViton alter the logo significantly, whereas GP-VTON unnaturally stretches the box on the shirt. ACDG-VTON (Ours) more faithfully represents the logo as the "EA7" looks the most similar to the garment input and the box lines are parallel. The "Emporio Armani" text is still illegible, however, as 512x384 is too low resolution. While HR-Zoom represents the text better, the text is too small for the VAE to properly encode and decode.

In row 2, Ladi-VTON unnaturally cuts the garment off, StableViton adds sleeves to the sleeveless top, and GP-VTON removes the stitching from the front and also generates an unnaturally large arm with artifacts at the elbow. In contrast, ACDG-VTON (Ours) includes the stitching in the front of the garment and correctly displays the garment as a sleeveless top. Additionally, in HR Zoom, our result shows the garment flared out at the hem, correctly indicating a looser fit.

In row 3, Ladi-VTON and StableViton alter the text completely. GP-VTON faithfully reconstructs the text, but the text is a bit grainy. In contrast, Ours and HR-Zoom both display the text clearly and legibly.

In row 4, Ladi-VTON and StableViton completely alter the blue crab logo and add long sleeves to the t-shirt. GP-VTON preserves the logo, but has sleeve artifacts on the arms and at the hip. Ours and HR-Zoom faithfully reconstruct the crab logo and correctly generate the short sleeve shape of the shirt.

In row 5, Ladi-VTON and StableViton alters the flower patterns. Stable Viton also adds sleeves to the sleeveless top. GP-VTON faithfully reconstructs the textures, but makes the top a tailored fit, whereas the blouse looks like it should be a more natural fit. In contrast, Ours provides a more correct natural fit, but some fine details are slightly altered due to the VAE. However, HR-Zoom can faithfully reconstruct the flower patterns and maintain the natural fit of the top.

In row 6, Ladi-VTON adds white colors and folds to the sleeves that are not present in the garment image. StableViton changes the hue of the blue and makes the sleeve a bit too long. GP-VTON makes the sleeve too short and turned the blouse into a dress. In contrast, Ours and HR-Zoom generate the correct color and sleeve length.

In row 7, Ladi-VTON and StableViton alter the pattern on the top. GP-VTON adds dark splotches, making the garment look completely different. In contrast, Ours and HR-Zoom faithfully reconstruct the diamond patterns without any artifacts and can accurately capture the fuzzy texture of the top. 

\section{More Ablation Results}
\begin{figure}[tb]
  \centering
    \includegraphics[width=1.0\textwidth]{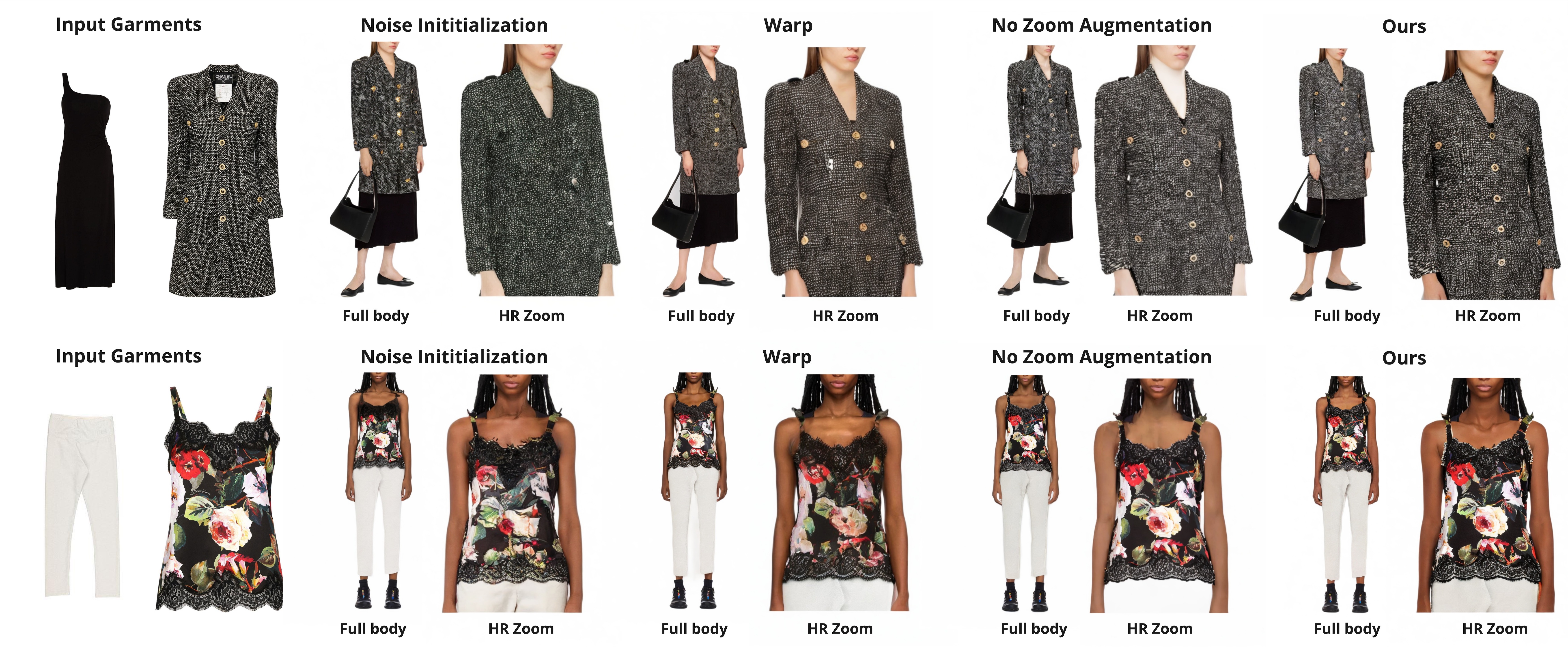}
  \caption{We show the effect of each training component on more examples. Notice noise initialization and warp show garment features being altered (buttons changed/missing, flowers and leaves distorted, etc.). Additionally, "No Zoom Augmentation" HR Zoom results have artifacts on the skin for both rows and the white pants on the bottom row do not render.}
  \label{fig:more_ablations}
\end{figure}

We visualize more ablation results comparing the impact of each component of our method in Fig.~\ref{fig:more_ablations}. The analysis from the main text is reflected in the additional figure.

For "Noise Initialization" results, we see severe alterations of garment details. In the top row, the buttons are not placed in the same positions as the input, and the HR Zoom generation has no buttons at all. In the bottom row, the features flowers and the leaves are significantly altered. 

For "Warp" results, we also see distortions and changes to the garment features. In the top row, the button shape is much larger in the full-body example, with some buttons removed. In the HR Zoom result, while the number of buttons is correct, the shape and look of the buttons are not the same as the input. In the bottom row, the flowers and leaves are also altered.

For "No Zoom Augmentation" results, we see full-body results represent the garments faithfully, but HR Zoom shows artifacts in the skin. In the top row for HR Zoom, the neck, chest, and hands are very pale. In the bottom row for HR Zoom, the skin looks less realistic, and the white pants have disappeared. 

\section{Failure Results}
\begin{figure}[tb]
  \centering
    \includegraphics[width=1.0\textwidth]{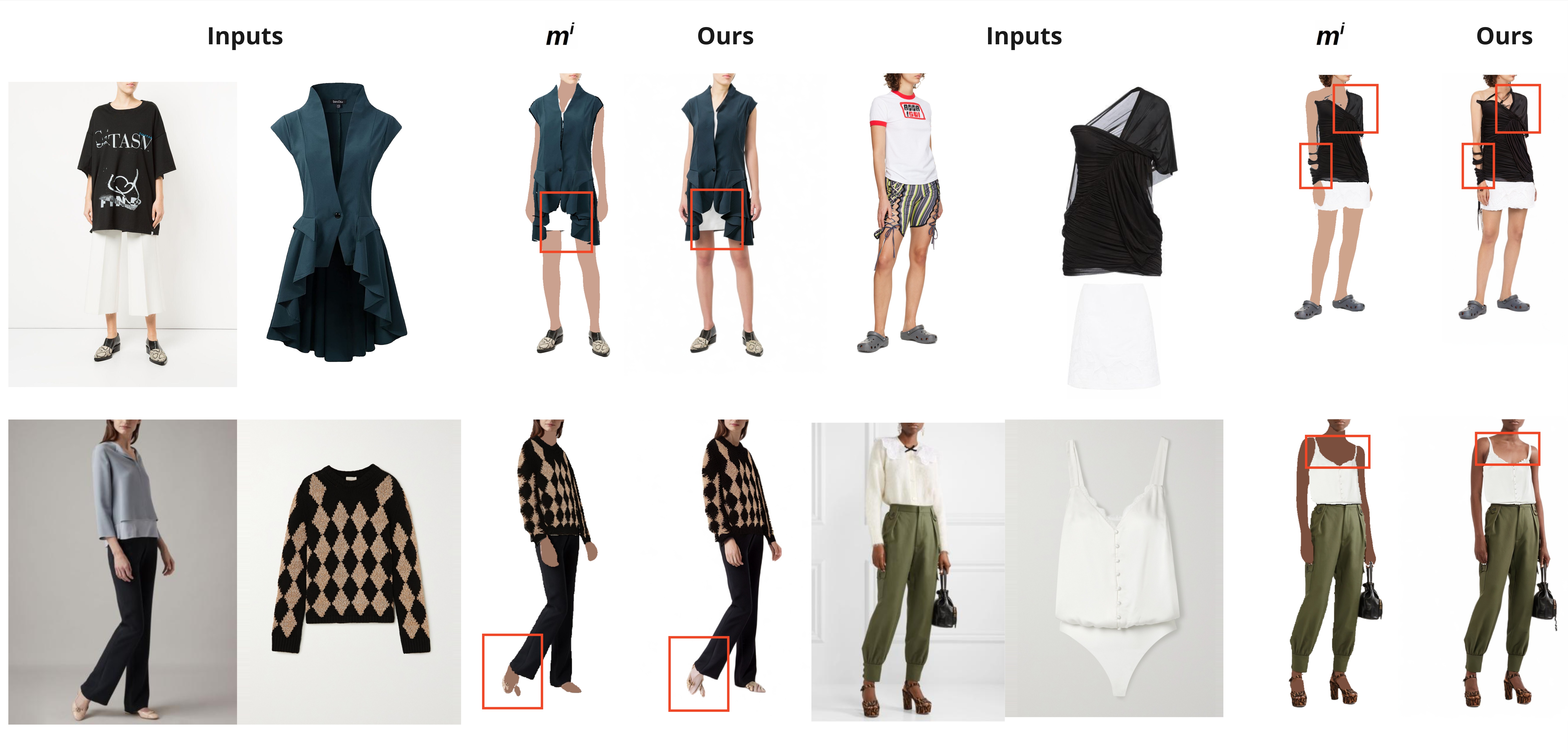}
  \caption{We show failure cases that our denoiser cannot fix. On the top left, we show that our processing method removes the back of garments, and these may not show properly for dresses or outerwear that shows the back of the garment between the legs. On the top right, we show that the transparency does not show correctly for the sleeve and the warper made some errors on the sleeveless region, causing our generation to mess up that region. On the bottom left, we notice the bad layout prediction around the foot, leading to distorted feet generation. Finally, on the bottom right, we see that the width of the strap is inaccurately warped, leading the generation to be inaccurate as well. }
  \label{fig:failures}
\end{figure}

Our method has some failures related to warping errors, specific garment types, or layout-related errors. In Fig.~\ref{fig:failures}, we show examples of failure modes. While our generator can rectify small errors, as seen in Fig.~\ref{fig:rendering_comparison_supplementary} and Fig.7 of the main text, some errors are more challenging to fix. On the top left example in Fig.~\ref{fig:failures}, we show that our processing method removes the back of garments, leading to incorrect results for dresses or outerwear that should show the back of the garment behind the legs. On the top right, we show that the transparency does not show correctly for the sleeve and the warper made some errors on the sleeveless region, causing our generation to generate artifacts on the sleeve and the chest. On the bottom left, we show errors that could occur from bad layout predictions. In this example, the bad layout prediction around the foot leads to distorted feet. Finally, on the bottom right, we see that the width of the strap can be inaccurately warped, leading to slight inaccuracies with the strap width in the generation.

One reason for these errors is that our training procedure does not encounter them often and has not learned to correct them. Adding these distortions as a form of augmentation could help teach the denoiser $F$ to fix some of these errors. Another reason is we do not feed in the unwarped garment into $F$. This makes it more difficult for the denoiser to infer how to fix potential issues as it does not have the original garment as a reference. Adding the unwarped garment as an additional control image could be another future improvement.

\end{document}